\documentclass[runningheads]{llncs}

\usepackage{eccv}

\usepackage{eccvabbrv}
\usepackage[accsupp]{axessibility}  
\usepackage{graphicx}
\usepackage{booktabs}
\usepackage{wrapfig}
\usepackage[numbers]{natbib}
\usepackage{url}            
\usepackage{amsfonts}       
\usepackage{nicefrac}       
\usepackage{microtype}      
\usepackage{xcolor}         
\usepackage{xspace}
\usepackage{graphicx}
\usepackage{booktabs}       
\usepackage{titletoc}
\usepackage{lipsum}
\usepackage[most]{tcolorbox}
\tcbset{
  finding style/.style={
    enhanced,
    colback=gray!10,           
    colframe=gray!60,         
    boxrule=0.8pt,             
    arc=2mm,                   
    left=1mm, right=1mm,       
    top=1mm, bottom=1mm,
    before skip=10pt, after skip=10pt,
    fontupper=\normalfont,     
    fonttitle=\normalfont\bfseries, 
  }
}
\newtcolorbox{findingbox}[1][]{finding style,#1}

\usepackage{hyperref}

\usepackage{orcidlink}
\newcommand\ours{\textsc{V-REX}\xspace}

\begin{document}

\title{\textsc{\textcolor{Cyan}{V}-\textcolor{Dandelion}{R}\textcolor{RedOrange}{E}\textcolor{LimeGreen}{X}\xspace}: Benchmarking Exploratory Visual Reasoning via Chain-of-Questions} 


\titlerunning{\textsc{\textcolor{Cyan}{V}-\textcolor{Dandelion}{R}\textcolor{RedOrange}{E}\textcolor{LimeGreen}{X}\xspace}: Benchmarking Exploratory Visual Reasoning via Chain-of-Questions
}


\author{\textbf{Chenrui Fan}$^{\star,1}$\textbf{, Yijun Liang}$^{\star,1}$\textbf{, Shweta Bhardwaj}$^{\star,1}$\\
\textbf{Ming Li}$^{1}$\textbf{, Kwesi Cobbina}$^{1}$\textbf{, Tianyi Zhou}$^{2}$ \\
{\tt\small {\{cfan42,yliang17,shweta12\}}@umd.edu}\\
\small{Project: \url{https://github.com/tianyi-lab/VREX}}\\
$^{\star}$These authors contributed equally to this work.
}
\institute{$^{1}$University of Maryland, College Park \quad $^{2}$Mohamed bin Zayed University of Artificial Intelligence}

\authorrunning{F.~Fan et al.}

\maketitle
\begin{abstract}
  While many vision language models (VLMs) are developed to passively answer well-defined, straightforward questions with highly specified targets, as in most benchmarks, they often struggle in practice with complex open-ended tasks, which usually require multiple rounds of active exploration and reasoning in the visual space. Such visual thinking paths not only provide step-by-step exploration and verification as an AI detective but also produce better interpretations of the final answers. However, these paths are challenging to evaluate due to the large exploration space of intermediate steps. 
  To bridge the gap, we develop an evaluation suite, ``\textbf{V}isual \textbf{R}easoning with multi-step \textbf{EX}ploration (\ours)'', which is composed of a benchmark of challenging visual reasoning tasks requiring native multi-step exploration and an evaluation protocol.
  \ours casts the multi-step exploratory reasoning into a Chain-of-Questions (CoQ) and disentangles VLMs' capability to 
  (1) Planning: determining the exploratory direction to gather information at each step for solving the task;
  and (2) Following: answering curated CoQ sequentially to collect information for deriving the final answer. 
  By curating finite options of questions and answers per step, \ours achieves a reliable quantitative and fine-grained analysis of the intermediate steps. 
  By assessing SOTA proprietary and open-sourced VLMs, we reveal consistent scaling trends, significant differences between planning and following abilities, and substantial room for improvement in multi-step exploratory reasoning.\looseness-1
\end{abstract}    
\section{Introduction}
\label{sec:intro}

Various practical applications of vision language models (VLMs) need to perform sophisticated multi-step visual reasoning \citep{idealgpt_emnlp2023, llavacot_iccv2025, multistep_neurips25, liu2024imagebasedgeolocationusinglarge, li2025visualtextgroundingmultimodal, chen2025visrbenchempiricalstudyvisual, li2025perceptionreasonthinkplan,wang2025multimodalchainofthoughtreasoningcomprehensive}
to solve the user queries. 
Recent studies reveal the weakness of existing VLMs on exploratory reasoning tasks, showing that they often rely on brute-force search in the input image to allocate the potential objects of interest, and rarely adjust their plans to be adaptive to collected clues~\cite{limitVLMs_neurips24, capture_iccv2025, li2025caughtcheatingmllmgoodcheating}. This weakness substantially limits the application of VLMs in challenging open environments where the goals cannot be fully specified at the very beginning but need progressive planning on the fly. 
For example, guessing the location based on a street view image~\cite{liu2024imagebasedgeolocationusinglarge}, detecting cheating from posted images~\cite{li2025caughtcheatingmllmgoodcheating}, or simply complicated tasks~\cite{roberts2025zerobenchimpossiblevisualbenchmark}.  
Tasks like these require multiple rounds of active exploration, sub-goal proposal, and answering the sub-questions to collect sufficient contextual clues and identify the final targets, while poor exploration may significantly undermine or distract such reasoning processes. \looseness-1

\begin{wrapfigure}{r}{0.5\textwidth}
\centering
\vspace{-15pt}
\includegraphics[width=\linewidth]{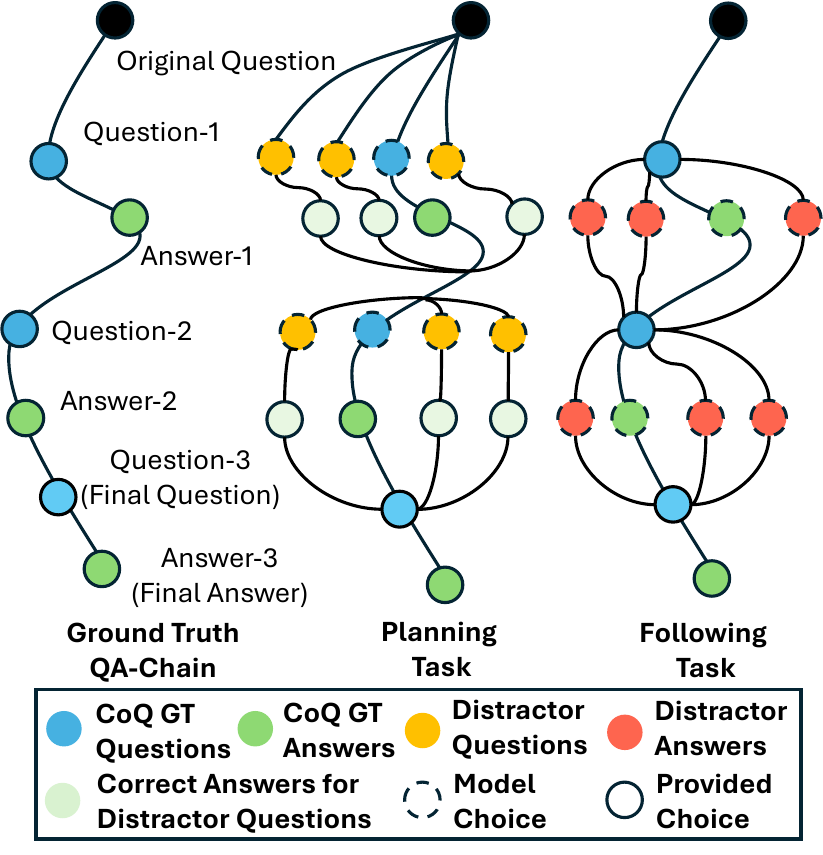}
\vspace{-4mm}
\caption{\textbf{Overview of Chain-of-Questions (CoQ)}. The left represents the manually formulated ground truth QA chain. The middle represents the \textit{Planning} task, evaluating the model's capability in selecting sub-questions that are helpful to answer the original question. The right represents the \textit{Following} task, evaluating the model's capability in answering each sub-question. }
\label{fig:intro}
\vspace{-4mm}
\end{wrapfigure}

\begin{figure*}[t]
\centering
\includegraphics[width=\linewidth]{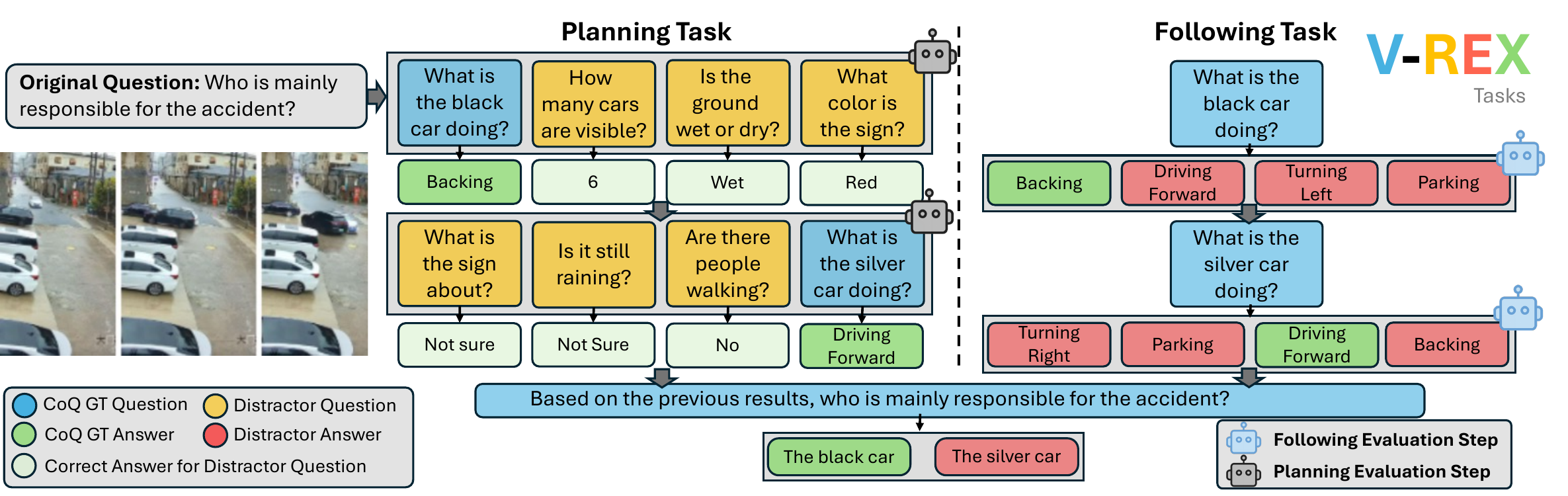}
\caption{\textbf{An example from \ours with corresponding \textit{planning} and \textit{following} tasks.} 
In the \textit{planning} task, the model is given the original question and asked to select a sub-question in each step that is necessary and helpful for solving the original question. 
In the \textit{following} task, the model is asked to answer the ground truth sub-questions step-by-step. 
} 
\label{fig:curation}
\vspace{-4mm}
\end{figure*}

However, recent visual reasoning and benchmarks mainly focus on math problems with visual contexts~\cite{xu2025visulogicbenchmarkevaluatingvisual, lu2024mathvistaevaluatingmathematicalreasoning, wang2025benchmarkingmultimodalmathematicalreasoning} or puzzle games with toy environments~\cite{ren2025vgrpbenchvisualgridreasoning, lyu2025jigsawpuzzlesseeingunderstandingreasoning, song2025visualpuzzlesdecouplingmultimodalreasoning}, which can be addressed by single-round QA or language space reasoning. These questions are usually straightforward with highly specified instructions or targets. In addition, their input prompt and context are sufficient for the models to make a concrete plan in advance and execute it without further exploration. 
Hence, they do not heavily rely on exploration skills, which, in contrast, are critical to open-ended tasks that need to actively collect clues given only high-level initial intents. 
Moreover, most of them merely evaluate the final answers and do not investigate the intermediate exploration steps, which can reflect important capabilities or fundamental flaws, such as shortcut solutions or redundancy, making the whole evaluation suite a black box. 
Hence, it is still unclear how each step affects the subsequent reasoning and the final answer, how robust existing VLMs are to intermediate-step errors, and whether they can recover from these errors. \looseness-1

\begin{figure*}[t]
\centering
\includegraphics[width=\linewidth]{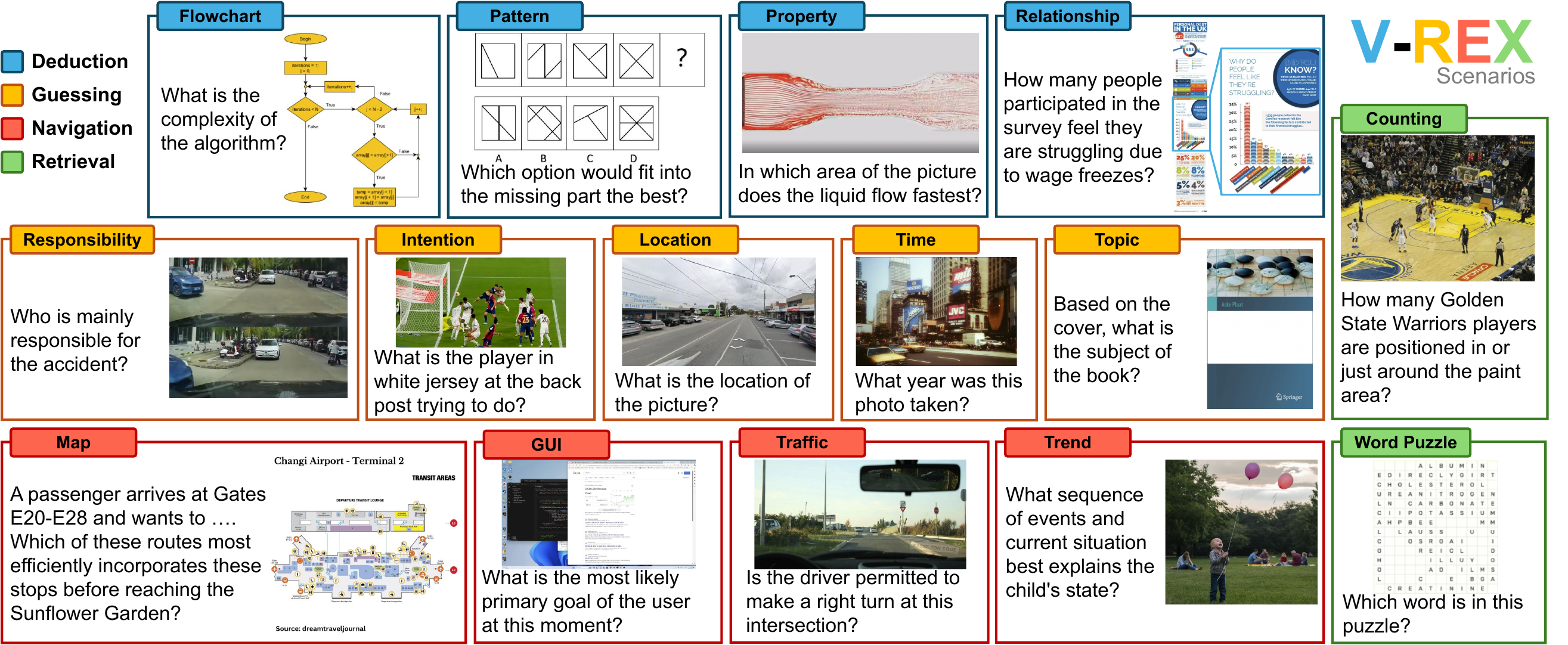}
\caption{\textbf{Scenarios in \ours}. 
With various samples, \ours spans $15$ real-world scenarios across $4$ reasoning categories (\textit{Deduction}, \textit{Guessing}, \textit{Navigation}, \& \textit{Retrieval}), covering diverse settings such as diagrams, time estimation, GUI interpretation, and others.}
\label{fig:vrex_scenarios}
\vspace{-4mm}
\end{figure*}

To bridge the gap, we investigate \textit{the quality and impact of the intermediate exploration steps} and develop a novel evaluation suite, ``\textbf{V}isual \textbf{R}easoning with multi-step \textbf{EX}ploration (\textbf{\ours})'', which is composed of challenging visual reasoning tasks requiring native multi-step exploration and subgoal decomposition. 
Specifically, we formulate the exploratory reasoning process in a novel form called ``\textbf{Chain-of-Questions (CoQ)}'', 
which represents a sequence of interconnected sub-questions and corresponding answers that serve as the precondition for the final answer, as shown in Figure \ref{fig:intro} (left). 
In CoQ, a model is prompted to ask a sequence of sub-questions and answer them sequentially to collect sufficient context, thereby creating a \textbf{QA chain} progressively on the fly to address the original question. It is worth clarifying that CoQ is 
\textbf{not} a prompting method(e.g. CoT) to improve models' performance, but a way to conduct fine-grained evaluation of the exploratory visual reasoning ability of model.


A key challenge of evaluating multi-step exploratory reasoning is the prohibitively large space for free exploration of VLMs. 
CoQ can reduce it to a finite space with limited choices of questions and answers per step, providing a tractable and controllable assessment. 
Specifically, CoQ enables us to disentangle the reasoning capabilities into \textbf{\textit{Planning}} and \textbf{\textit{Following}} when facing complex tasks that require multiple rounds of exploration with sub-goal proposals. 
The \textit{Planning} of CoQ evaluates the model’s exploratory ability to determine effective directions by selecting informative subquestions that guide information gathering toward the final answer.
For example, in Figure~\ref{fig:curation} (middle), when given the question ``Who is mainly responsible for the accident?'', the model should raise exploratory questions about the conditions of the two vehicles, rather than questions about the number of vehicles. 
The \textit{Following} of CoQ evaluates the model's ability to answer the subquestions to collect sufficient contextual clues and identify the final targets, as shown in Figure \ref{fig:curation} (right). 
An overview of CoQ is shown in Figure \ref{fig:intro}, where the two dimensions serve as the complementary parts of the exploratory reasoning process, indicating a holistic and fine-grained diagnosis of the reasoning capabilities of VLMs.\looseness-1

\ours comprises $702$ samples and $2,504$ questions spanning $4$ reasoning categories and $15$ scenarios as shown in Figure \ref{fig:vrex_scenarios}. 
Each sample contains $2$ to $6$ reasoning steps, with $3.57$ steps on average. 
In our experiments, we extensively evaluate VLMs of different model families, ranging from open-source to proprietary models, including relatively small models and larger ones. The evaluation lead to some unrevealed observations and shed novel insights to improving the exploratory reasoning capability of VLMs.

\textbf{Contributions:} 
We introduce \ours, the first benchmark for assessing the multi-step exploratory reasoning capability of VLMs with a novel form of Chain-of-Questions (CoQ). 
In our setting, the model's planning and following abilities can be disentangled and evaluated separately, providing a more fine-grained and interpretable evaluation of the reasoning capabilities of VLMs. 

\textbf{Key Findings:} 
\begin{enumerate}
    \item By following the hints of CoQ, VLMs consistently achieve better performance on final questions, demonstrating the importance of exploration in visual reasoning. 
    \item The scaling law holds for \ours, and models of the same size show much less variance in \textit{Following} performance than in \textit{Planning} performance; this indicates a need to improve VLMs' exploratory abilities.
    \item \textit{Following} and \textit{Planning} capabilities both contribute positively to the overall performance of the model.  
    \item Smaller models are better at \textit{Following} than \textit{Planning} while larger models have more balanced performance.
    \item VLMs are better at recovering from failed \textit{Planning} than from failed \textit{Following}. 
\end{enumerate}

\section{Related Work}
\label{sec:related}


\subsection{Exploratory Visual Reasoning}

Inspired by recent advances on the reasoning ability of LLMs~\cite{deepseekai2025deepseekr1incentivizingreasoningcapability,openai2024openaio1card}, efforts have been made to replicate similar success on the multimodal domain. \cite{zhou2025r1zerosahamomentvisual, tan2025reasonrft, shen2025vlmr1stablegeneralizabler1style} adopt GRPO algorithm to train R1-style VLMs. \cite{Dong_2025_CVPR} designs an effective training pipeline to elicit VLM's ability to explore in visual space and general long and complex chain-of-thought.
Meanwhile, multiple benchmarks have been proposed to evaluate the visual reasoning ability of VLMs. VisuLogic~\cite{xu2025visulogicbenchmarkevaluatingvisual} measures the ability of VLMs on genuine vision-centric reasoning tasks. 
ZeroBench~\cite{roberts2025zerobenchimpossiblevisualbenchmark} evaluates VLMs on extremely difficult tasks. 
CaughtCheating~\cite{li2025caughtcheatingmllmgoodcheating} investigates the VLM's capability on cheating detection tasks, revealing the weakness of existing models on real detective-level tasks. 
More benchmarks evaluate VLMs' visual reasoning abilities from different aspects, including cognitive reasoning~\cite{pandya-etal-2025-ntsebench}, visual grid reasoning~\cite{ren2025vgrpbenchvisualgridreasoning}, visual comparison reasoning~\cite{cai2025comparebenchbenchmarkvisualcomparison}, cultural reasoning~\cite{satar-etal-2025-seeing}, multi-image reasoning~\cite{cheng2025evaluatingmllmsmultimodalmultiimage}, and color-related reasoning~\cite{liang2025colorbenchvlmsunderstandcolorful}. However, although these benchmarks cover a wide range of visual reasoning tasks, they mainly focus on the exploration skills of VLMs on answer space and neglect the exploration of question space, that is, the model's ability to propose helpful sub-questions, set subgoals that would guide the reasoning process effectively. Our benchmark, \ours, fills the gap by evaluating both exploration of answer space and question space in a disentangled manner.



\subsection{Evaluation with Intermediate Steps}
Most visual reasoning benchmarks~\cite{roberts2025zerobenchimpossiblevisualbenchmark, xu2025visulogicbenchmarkevaluatingvisual, pandya-etal-2025-ntsebench, cheng2025evaluatingmllmsmultimodalmultiimage} evaluate the ability of VLMs in end-to-end approach. 
Given complex visual reasoning tasks, the models are prompted to directly produce the final answer, and performance is measured solely by answer accuracy. 
Chen et al. \cite{chen2025unveilingchainstepreasoning} introduces a process reward model (PRM) to score intermediate reasoning steps. 
But they rely on Monte-Carlo estimation and LLM-as-a-Judge as PRM for intermediate steps, which are not reliable for benchmark evaluation.
LlamaV-o1~\cite{thawakar-etal-2025-llamav} adopts a fine-grained evaluation framework that compares each generated reasoning step to a ground-truth trace, but it also relies on LLM-based judgments when assessing alignment or correctness. 
By breaking down the complex visual reasoning tasks into multiple sub-tasks and evaluating the intermediate steps, \ours can more accurately assess the ability of VLMs to reason step-by-step in \textit{Planning} and \textit{Following} dimensions. 










\section{Exploratory Reasoning in Visual Space}
\label{sec:formulation}



\subsection{Chain-of-Questions (CoQ)}
To investigate the intermediate exploratory reasoning processes of vision language models (VLMs), we conceptualize their reasoning pathways in the CoQ format as a QA chain with a sequence of interconnected sub-questions $Q_t$ and corresponding answers $A_{Q_t}$. 
The reasoning process can then be denoted as: $\{(Q_1, A_{Q_1}), (Q_2, A_{Q_2}), ..., (Q_T, A_{Q_T})\}$, where $T$ is the number of reasoning steps. 
This formulation helps formulate the model's exploratory reasoning process as an iterative alternation between generating exploratory sub-questions and providing corresponding answers. 
It allows us to decouple the model's reasoning capability into two fundamental dimensions: the proficiency in formulating informative intermediate questions (\textit{Planning}), and the accuracy in addressing these sub-questions (\textit{Following}). 
However, evaluating these abilities in the infinite exploratory space is not feasible. It is challenging to design a reliable evaluation protocol because it involves directly rating the quality of open-ended reasoning processes. To address this, we propose to evaluate the model's ability to answer a series of multiple-choice questions (MCQ) with the CoQ process, reformulating the evaluation into a more tractable and controllable assessment in finite space. \looseness-1

\subsection{Planning of CoQ}
An important dimension of exploratory reasoning capability is \textit{Planning}: the capacity to strategically chart a course of exploration through the reasoning space. 
Effective planning entails the model's ability to dynamically adapt its sequence of inquiry based on evolving intermediate evidence and to accurately select the most informative next step at each stage. 
As illustrated in the middle of Figure~\ref{fig:curation}, we branch over the question node at each step of the pre-defined ground-truth QA chain by adding several uninformative, distracting questions. 
The planning evaluation then becomes asking the model $p$ to identify the helpful sub-question at step $t$ conditioned on the previous CoQ history:
{\small
\begin{equation}
    Q_t \sim p(\cdot|Q_{\text{origin}}, Q_1, A_{Q_1}^{*}, Q_2, A_{Q_2}^{*}, \ldots, Q_{t-1}, A_{Q_{t-1}}^{*})
\end{equation}
}
where $Q_{\text{origin}}$ is the original problem, $Q_t$ is the question selected by model at step $t$, and $A_{Q_t}^{*}$ is the ground-truth answer for $Q_t$. For simplicity, we omit the notation of the image in the above equation. 
By directly providing the corresponding ground-truth answer $A_{Q_t}^{*}$ to the model at each step, we ensure that the model does not need to solve the sub-questions itself but only needs to decide which question to ask next. 
This design explicitly isolates the \textit{Planning} ability from \textit{Following}. \looseness-1



\subsection{Following of CoQ}
For the \textit{Following} dimension, our objective is to evaluate the model's ability to accurately answer intermediate questions based on both visual cues from different regions of the image and knowledge obtained through preceding reasoning steps. 
Specifically, we present the VLM with a pre-constructed, sequential chain of QA steps and require the model to provide answers at each stage en route to the final conclusion. As depicted in the right part of Figure~\ref{fig:curation}, we introduce distractor options at each answer node, transforming each step into a multiple-choice setting. Specifically, at reasoning step $t$ in the chain, we augment the correct answer with several incorrect alternatives. The model $p$ is tasked with selecting the correct answer to question $Q_t^*$ at $t$ step, conditioned on the image and the accumulated conversation history:
{\small
\begin{equation}
    A_{Q_t^*} \sim p(\cdot| Q_1^*, A_{Q_1^*}, Q_2^*, A_{Q_2^*}, \ldots, Q_{t-1}^*, A_{Q_{t-1}^*}, Q_t^*)
\end{equation}
}
where $Q_t^*$ is the ground-truth question at each step and $A_{Q_t^*}$ is the corresponding model-generated response. 
This methodology enables a principled evaluation of the model’s capacity to faithfully follow an intended reasoning path.


\section{\ours Benchmark}
\label{sec:overview}

\begin{figure}[t]
\centering
\includegraphics[width=0.9\linewidth]{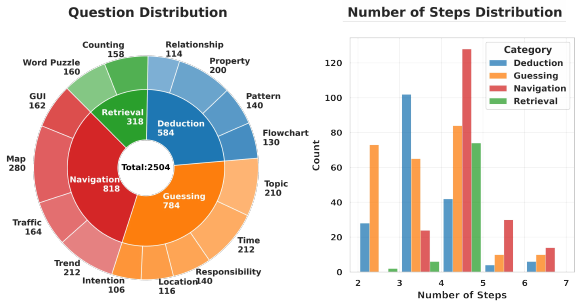}
\caption{\textbf{Statistics of \ours}, including question distributions (left) and the distribution of reasoning steps (right).
}
\label{fig:statistics}
\vspace{-4mm}
\end{figure}

\ours\ consists of $702$ samples and $2,504$ questions, with each sample containing $2$ to $6$ reasoning steps and $3.57$ steps on average. 
The detailed distributions of questions and reasoning steps are illustrated in Figure~\ref{fig:statistics}. 
The dataset spans $4$ reasoning categories and $15$ application scenarios, supporting a fine-grained evaluation of VLMs’ exploratory reasoning capabilities. 
\looseness-1

\subsection{Taxonomy}

To thoroughly evaluate exploratory reasoning, \ours defines $4$ reasoning categories where exploration plays a central role: \textbf{{Deduction}}, \textbf{{Guessing}}, \textbf{{Navigation}}, and \textbf{{Retrieval}}. 
Each category reflects a different way VLMs explore and reason across multi-step problems.

\textbf{Deduction} category measures model's ability to discover, infer, and apply logical or causal relationships from structured information. 
It includes $4$ application scenarios: 
(i) \textit{Flowchart}, 
(ii) \textit{Pattern}, 
(iii) \textit{Property}, and 
(iv) \textit{Relationship}. 
\textbf{Guessing} category involves uncertainty and incomplete information, requiring models to infer hidden or missing factors through hypothesis exploration. 
It includes $5$ scenarios: 
(i) \textit{Responsibility}, 
(ii) \textit{Intention}, 
(iii) \textit{Location}, 
(iv) \textit{Time}, and 
(v) \textit{Topic}. 
\textbf{Navigation} category measures the model’s ability to explore by traversing spatial layouts or procedural paths, planning step-wise movements through the scene while maintaining global consistency.
It includes $4$ scenarios: 
(i) \textit{Map}, 
(ii) \textit{GUI}, 
(iii) \textit{Traffic}, and 
(iv) \textit{Trend}. 
\textbf{Retrieval} category measures the model’s ability to locate, gather dispersed information by thoroughly exploring the visual input.
It includes $2$ scenarios: 
(i) \textit{Counting} and 
(ii) \textit{Word Puzzle}. 

\subsection{Data Curation}
We begin by elaborating on the methodology for constructing the ground-truth QA chains for each sample, followed by an explanation of how these chains are systematically adapted to probe the distinct following and planning capabilities of VLMs within the reasoning space. 

\subsubsection{Ground-Truth QA Chain}
As discussed in Section~\ref{sec:formulation}, we use QA chains to model the reasoning traces of VLMs when solving multi-step reasoning problems. 
We denote the ground-truth QA chain as a sequence $\{(Q_1^*,A_{Q_1^*}^*), (Q_2^*,A_{Q_2^*}^*), ..., (Q_T^*,A_{Q_T^*}^*)\}$, where $Q_T^*$ is the final question and $A_{Q_T^*}^*$ is the final answer, a valid QA chain should have the following properties:
\begin{findingbox}
\begin{itemize}
    \item \textbf{Helpfulness.} $\forall t\in[1, T-1]$, $\exists t'>t$,  $A_{Q_t^*}^*$ is helpful for answering $Q_{t_2}^*$.  It means that every question should be helpful for the following questions. 
    \item \textbf{Correctly Ordered.} $\forall t\in[2, T]$, $\forall t'<t$, $Q_{t'}^*$ does not depend on $A_{Q_t^*}^*$. It means that every question should not depend on its subsequent question. 
\end{itemize}
\end{findingbox}

We rely on human experts to collect high-quality ground-truth QA chains on visual reasoning tasks across all categories. 
The images are sourced from websites and publicly available benchmarks, with detailed sources provided in Appendix~\ref{appendix:data_sources}.
Specifically, five PhD-level annotators are initially assigned three scenarios each for ground-truth QA chain construction. 
We then have two rounds of cross-verification. 
For each round, annotators are randomly shuffled to three other unseen scenarios. 
They are required to ensure the validity of the reasoning chain, helpfulness of the intermediate steps, and make sure they are in the correct order. 
At the end of each verification round, they would give feedback to the original annotator and the previous verifier.

\subsubsection{Planning Task}
To evaluate VLMs’ exploratory ability in the question space, we construct samples for the \textit{Planning} dimension based on ground-truth QA chains.
We develop a two-stage automatic generation pipeline, consisting of \textbf{Distractor Construction} and \textbf{Question Integration}, which leverages LLM for generation, filtering, and integration. 

\textbf{Distractor Construction} stage aims to create alternative questions that are contextually relevant but unhelpful for solving the final question. 
We design two complementary strategies to capture local and global variability in reasoning: \textit{{step-level}} and \textit{{chain-level}} distractor generation. 
In {{step-level}} generation, LLM (GPT-5) produces distractors for each reasoning step that are locally plausible but deliberately misleading. 
In contrast, {{chain-level}} generation considers the entire distracting QA chain holistically. 
These chains are self-consistent yet subtly deviate from the ground truth, allowing evaluation of whether models can distinguish globally misleading reasoning paths from the true reasoning chain. 
\textbf{Question Integration} stage aims to integrate the generated distractors with the ground-truth chain into a unified dataset through automatic filtering and refinement, creating a controlled benchmark for evaluating models’ exploration reasoning in complex question spaces. 
More details are provided in the Appendix~\ref{appendix:curation_planning}.

\subsubsection{Following Task}
For \textit{Following} dimension, we manually create a few plausible distractor choices for each answer step. 
These distractors are combined with the ground-truth answer to form a series of multiple-choice questions.

\subsection{Evaluation Metrics}
We use \textit{intermediate} accuracy to measure both \textit{Following} and \textit{Planning} ability of the model. Specifically, for a QA chain of $T$ steps, \textit{Planning} ability is quantified as:
{\small
\begin{equation}
    \frac{1}{T-1} \sum_{t=1}^{T-1} \mathbb{I}[Q^*_t = Q_t]
\end{equation}
}
where $Q^*_t$ is the ground-truth question at step $t$ and $Q_t$ is the model's chosen question at step $t$. 
Similarly, \textit{Following} ability is quantified as:
{\small
\begin{equation}
    \frac{1}{T-1} \sum_{t=1}^{T-1} \mathbb{I}[A^*_{Q_t^*} = A_{Q_t^*}]
\end{equation}
}
where $\mathbb{I}[\cdot]$ is the indicator function, $A_{Q_t^*}$ is the model's chosen answer for the ground-truth question $Q_t^*$ at step $t$, and $A^*_{Q_t^*}$ is the ground-truth answer.


\section{Results}
\label{sec:results}
\subsection{Main Results}
\begin{table*}[t]
\centering
\caption{\textbf{Performance of various VLMs (grouped by size) on \ours.} The best performance in each VLM group is highlighted in \textbf{bold}. 
}
\label{tab:main-result}
\resizebox{\linewidth}{!}{\begin{tabular}{l|cccc|c|cccc|c}
\toprule
& \multicolumn{5}{c|}{\textbf{Planning}} & \multicolumn{5}{c}{\textbf{Following}} \\
\cmidrule(lr){2-6}\cmidrule(l){7-11}
\textbf{Model} & Deduction & Guessing & Navigation & Retrieval & \textbf{Average} & Deduction & Guessing & Navigation & Retrieval & \textbf{Average} \\
\midrule
\multicolumn{11}{c}{VLMs: $<$7B} \\
\midrule
LLaVA-OV-1B~\cite{li2024llavaonevisioneasyvisualtask} & 31.5 & 37.1 & 27.9 & 43.9 & 35.1 & 33.8 & 49.3 & 64.0 & 56.1 & 50.8 \\
InternVL3-1B~\cite{zhu2025internvl3} & 37.1 & 47.3 & 42.5 & 42.4 & 42.3 & 37.4 & 50.2 & 67.1 & 61.0 & 53.9 \\
InternVL3.5-1B~\cite{wang2025internvl3} & 42.4 & 43.5 & 42.2 & 41.1 & 42.3 & 44.9 & 45.2 & 69.8 & 62.2 & 55.5 \\
Qwen3-VL-2B-IT~\cite{yang2025qwen3} & 31.9 & 37.5 & 34.1 & 28.4 & 33.0 & 50.2 & 61.0 & 80.3 & 48.0 & 59.9 \\
Qwen3-VL-2B-Think~\cite{yang2025qwen3} & 44.4 & 60.6 & 48.4 & 41.9 & 48.8 & 39.2 & 51.6 & 72.9 & 48.4 & 53.0 \\
InternVL3-2B~\cite{zhu2025internvl3} & 52.7 & 53.8 & 64.1 & 36.9 & 51.9 & 51.6 & 60.8 & 80.7 & 61.0 & 63.5 \\
InternVL3.5-2B~\cite{wang2025internvl3} & 63.1 & 60.8 & 63.2 & 42.7 & 57.5 & 59.3 & 51.5 & 71.3 & 60.2 & 60.5 \\
Qwen2.5-VL-3B~\cite{bai2025qwen25vltechnicalreport} & 47.9 & 59.3 & 51.0 & 37.3 & 48.9 & 61.2 & 57.7 & \textbf{83.4} & 66.7 & 67.2 \\
Qwen3-VL-4B-IT~\cite{yang2025qwen3} & 62.5 & 65.4 & \textbf{70.9} & 36.8 & 58.9 & 55.7 & \textbf{63.8} & 81.6 & 56.9 & 64.5 \\
Qwen3-VL-4B-Think~\cite{yang2025qwen3} & 58.8 & \textbf{83.4} & 63.6 & 42.2 & 62.0 & 56.2 & 57.8 & 82.7 & 60.2 & 64.2 \\
InternVL2.5-4B~\cite{chen2024internvl} & 62.5 & 57.9 & 69.0 & \textbf{55.4} & 61.2 & 55.3 & 60.9 & 77.6 & 65.9 & 64.9 \\
InternVL3.5-4B~\cite{wang2025internvl3} & \textbf{69.6} & 65.4 & 66.8 & 51.8 & \textbf{63.4} & \textbf{63.7} & 58.7 & 81.6 & \textbf{72.4} & \textbf{69.1} \\
\midrule
\multicolumn{11}{c}{VLMs: 7B$-$10B} \\
\midrule
LLaVA-OV-7B~\cite{li2024llavaonevisioneasyvisualtask} & 53.0 & 42.7 & 61.0 & 35.4 & 48.0 & 58.6 & 62.1 & 71.9 & \textbf{74.0} & 66.6 \\
Qwen2.5-VL-7B~\cite{bai2025qwen25vltechnicalreport} & 59.1 & 67.6 & 64.7 & 36.7 & 57.1 & \textbf{65.8} & 64.5 & 81.0 & 73.2 & \textbf{71.1} \\
Qwen3-VL-8B-IT~\cite{yang2025qwen3} & \textbf{74.7} & 84.7 & 80.0 & 46.8 & 71.6 & 59.2 & 64.2 & \textbf{88.3} & 57.7 & 67.3 \\
Qwen3-VL-8B-Think~\cite{yang2025qwen3} & 66.8 & \textbf{86.7} & 69.7 & 48.7 & 68.0 & 63.0 & 60.3 & 84.9 & 58.5 & 66.7 \\
InternVL2.5-8B~\cite{chen2024internvl} & 70.1 & 75.1 & 71.8 & 50.6 & 66.9 & 57.9 & 59.9 & 81.4 & 66.7 & 66.5 \\
InternVL3-8B~\cite{zhu2025internvl3} & 53.5 & 44.5 & 66.8 & 40.1 & 51.2 & 54.9 & 58.8 & 81.6 & 69.1 & 66.1 \\
InternVL3.5-8B~\cite{wang2025internvl3} & 66.2 & 69.6 & 73.1 & 47.3 & 64.1 & 55.6 & 61.0 & 82.8 & 61.8 & 65.3 \\
InternVL3-9B~\cite{zhu2025internvl3} & 69.1 & 81.6 & \textbf{84.1} & \textbf{57.5} & \textbf{73.1} & 56.0 & \textbf{66.3} & 82.9 & 69.1 & 68.6 \\
\midrule
\multicolumn{11}{c}{VLMs: $>$10B} \\
\midrule
InternVL3-14B~\cite{zhu2025internvl3} & 76.4 & 86.4 & 79.8 & 56.7 & 74.8 & 68.8 & 65.9 & 85.7 & 69.1 & 72.4 \\
InternVL3.5-14B~\cite{wang2025internvl3} & 78.6 & 73.0 & 77.9 & 53.3 & 70.7 & 64.9 & 62.3 & 85.1 & 72.4 & 71.2 \\
InternVL2.5-26B~\cite{chen2024internvl} & 76.4 & 71.1 & 79.9 & 52.6 & 70.0 & 64.4 & 61.0 & 85.5 & 69.9 & 70.2 \\
InternVL2.5-38B~\cite{chen2024internvl} & \textbf{85.6} & 85.0 & 85.0 & 59.8 & 78.8 & \textbf{71.3} & \textbf{69.0} & 90.8 & \textbf{74.8} & \textbf{76.5} \\
InternVL3-38B~\cite{zhu2025internvl3} & 83.2 & \textbf{91.3} & \textbf{86.6} & \textbf{65.2} & \textbf{81.6} & 68.1 & 65.2 & \textbf{91.8} & 69.1 & 73.5 \\
InternVL3.5-38B~\cite{wang2025internvl3} & 81.7 & 85.3 & 84.3 & 58.0 & 77.3 & 65.9 & 63.0 & 84.1 & 64.2 & 69.3 \\
\midrule
\multicolumn{11}{c}{VLMs: Proprietary} \\
\midrule
GPT-4o~\cite{openai2024gpt4ocard} & 77.9 & 69.6 & 79.7 & \textbf{63.7} & 72.7 & 73.1 & 78.0 & 87.1 & 67.5 & 76.4 \\
GPT-5~\cite{openai2025gpt5} & 93.9 & 91.8 & \textbf{93.9} & 58.1 & 84.4 & 76.6 & \textbf{86.2} & \textbf{90.9} & \textbf{84.6} & \textbf{84.6} \\
O1~\cite{openai2024openaio1card} & 94.4 & 93.2 & 92.4 & 61.3 & 85.3 & 73.0 & 80.4 & 90.7 & 74.0 & 79.5 \\
O3~\cite{openai2025o3o4mini} & \textbf{94.8} & \textbf{94.4} & 93.3 & 62.1 & \textbf{86.1} & \textbf{80.2} & 83.0 & 89.9 & 83.7 & 84.2 \\
Gemini 2.0 Flash~\cite{deepmind_gemini_flash} & 86.2 & 93.4 & 85.9 & 54.6 & 80.0 & 74.3 & 67.6 & 88.7 & 78.9 & 77.4 \\
Gemini 2.5 Flash~\cite{comanici2025gemini} & 87.8 & 89.7 & 84.5 & 49.6 & 77.9 & 69.8 & 74.2 & 87.1 & 78.0 & 77.3 \\
\bottomrule
\end{tabular}}
\vspace{-2mm}
\end{table*}


Table~\ref{tab:main-result} summarizes VLM performance across the \textit{Following} and \textit{Planning} tasks, revealing consistent scaling trends and distinct strengths across model families. Here, the performance refers to \textit{Following} and \textit{Planning} abilities. 
Overall, model performance increases steadily with size. 
Larger models show stronger multimodal reasoning and coordination capabilities. 
GPT-5 and o3 achieve the highest average performance on both tasks. 
Notably, the performance gap between large open-sourced and proprietary models has narrowed considerably. 
Several large open-source VLMs match or even exceed proprietary models in some reasoning categories, particularly Deduction and Navigation.

While zooming into different reasoning categories, the result reveals that VLMs exhibit distinct areas of expertise across model scales. 
Smaller models ($<$7B and 7B$\text{–}$10B) show highly variable strengths, with some excelling in Deduction or Guessing and others in Navigation or Retrieval. 
As the model size increases beyond 10B, the expertise distribution becomes more concentrated. 
On the \textit{Following} task, InternVL2.5-38B achieves the best accuracy in $3$ of $4$ categories, and InternVL3-38B similarly dominates in $3$ categories on the \textit{Planning} task.
Proprietary models such as GPT-5 and o3 also achieve top accuracy across multiple categories. 
These patterns suggest that larger models develop more unified and generalizable reasoning capabilities, rather than relying on specialized strengths in isolated categories.

\subsection{Main Findings}

\begin{findingbox}[title={Finding 1}]
By following the hints of CoQ, VLMs consistently achieve better performance on final questions, demonstrating the importance of exploration in visual reasoning.
\end{findingbox}

\begin{figure}[t]
\centering
\includegraphics[width=0.8\linewidth]{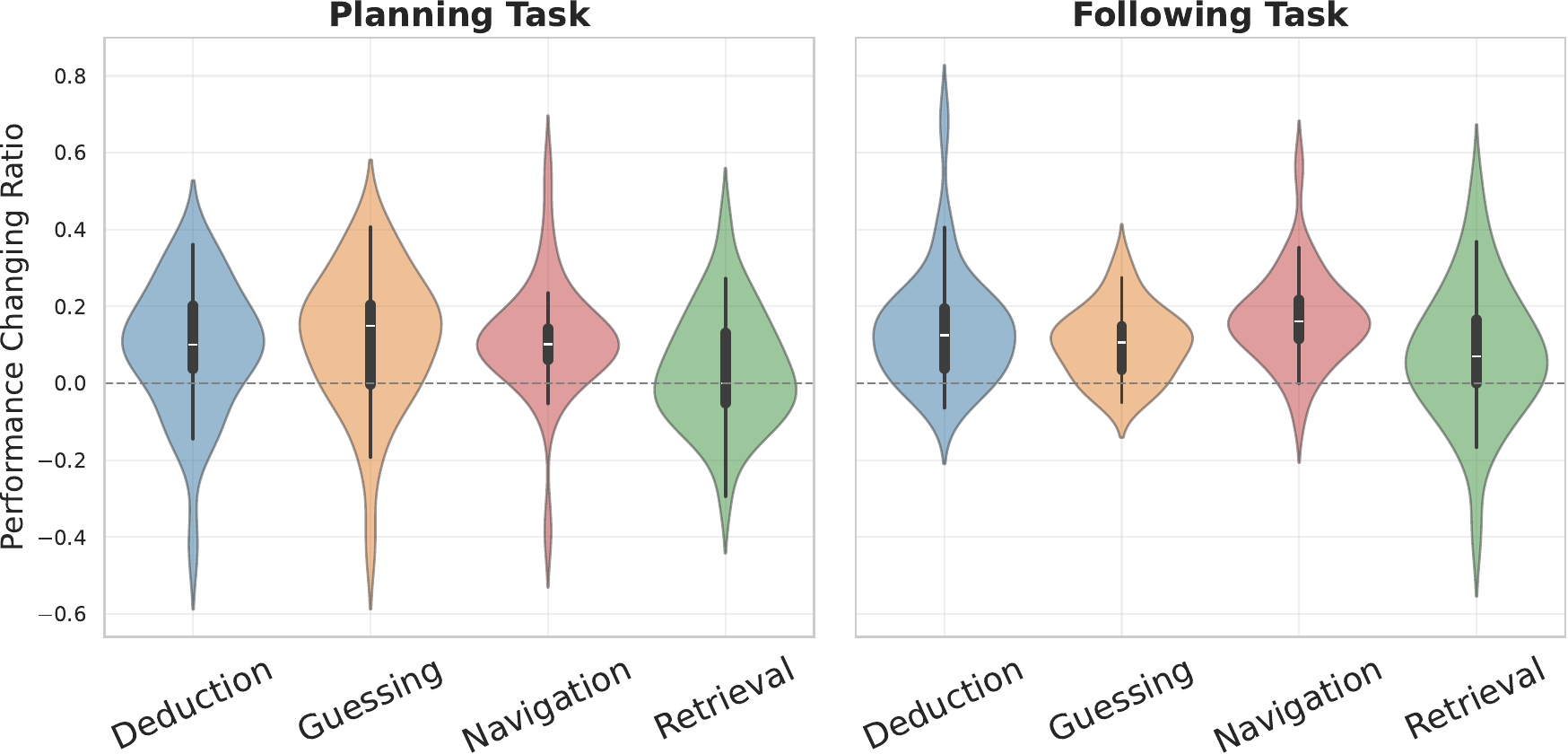}
\caption{
\textbf{The ratio of change on final accuracy brought by CoQ evaluation across all VLMs.} 
The X-axis denotes task categories, and the Y-axis represents the performance changing ratio.
\looseness-1}
\vspace{-4mm}
\label{fig:acc_diff}
\end{figure}

To assess the effect of the hints in CoQ in VLMs, we compare the accuracy performance on final questions with and without these hints across all models.
For quantitative analysis, we compute the performance changing ratio for each VLM and reasoning category as \((\text{Acc}_{CoQ} - \text{Acc})/\text{Acc}\). 
The detailed accuracies on $4$ categories for each VLM can be found in Appendix~\ref{appendix:final_acc}. 
Figure~\ref{fig:acc_diff} illustrates the distribution of performance changing ratios across all models for the \textit{Following} and \textit{Planning} tasks. 
Most ratios are positive, demonstrating that the model achieves higher accuracy on final questions when intermediate steps are introduced.
This indicates that the partial hints we provide to the models on both tasks (i.e., questions in \textit{Following} task and answers in \textit{Planning} task) help VLMs in reasoning, which justifies the validity of our manually constructed ground-truth QA chains. \looseness-1

The Retrieval category shows relatively minor improvements in both tasks, likely because it depends more on factual recall and direct visual matching than on complex reasoning.
Consequently, the CoQ process contributes less to tasks where success relies primarily on precise information retrieval rather than the hierarchical organization of reasoning processes. 
Another possible reason is that the human-designed reasoning paths may diverge from the model’s intrinsic retrieval strategy. 
This mismatch suggests that while hints in CoQ effectively help structures with deductive or sequential reasoning, it may not optimally support tasks dominated by information lookup or dense visual-textual association. \looseness-1

Moreover, when comparing the \textit{Following} and \textit{Planning} tasks, a few VLMs exhibit decreased performance when provided with CoQ hints across all categories. 
For some models, the introduction of intermediate steps can generate distractive reasoning paths or omit essential reasoning cues in \textit{Planning} task settings, ultimately disrupting their decision-making process. 
As a result, instead of improving reasoning consistency, adding exploration steps may introduce additional cognitive noise that hinders accurate question or answer selection.
The failure cases are shown in Appendix~\ref{appendix:cases_study}.
\looseness-1

\begin{findingbox}[title={Finding 2}]
Scaling laws persist on \ours, while performance variance among models of the same size is smaller in \textit{Following} than in \textit{Planning}.
\end{findingbox}

\begin{figure}[t]
\centering
\includegraphics[width=0.9\linewidth]{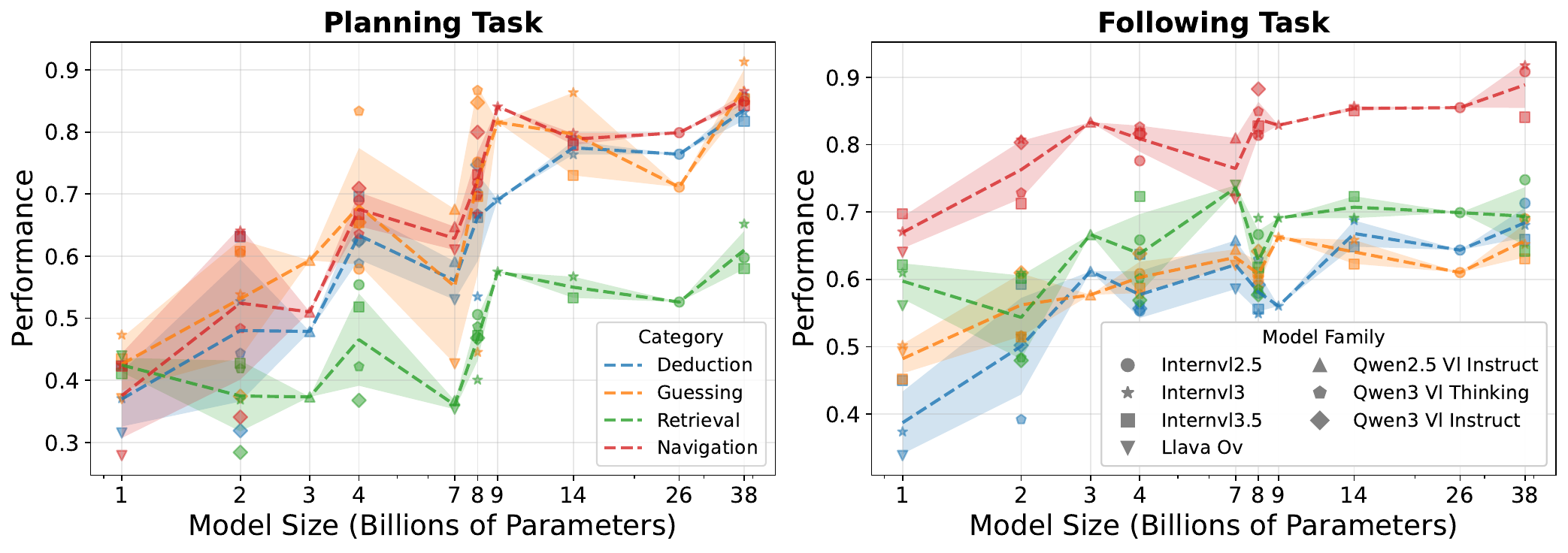}
\caption{\textit{Following} and \textit{Planning} ability on models of different sizes (logarithmic scale for the x-axis). 
\textbf{Overall, the model's ability on both tasks positively correlates with model size. Notably, the variance of \textit{following} ability among same-sized models is smaller than that of \textit{Planning} ability.}\looseness-1}
\label{fig:scaling_law}
\vspace{-4mm}
\end{figure}

To investigate whether the general scaling law holds on our \textit{Following} and \textit{Planning} tasks, we plot the model' abilities against parameter size for both tasks. 
As in Figure~\ref{fig:scaling_law}, colors denote the sample categories and marker shapes represent different model families. 
We also use dashed lines with shaded bands to aggregate the ability of models of the same scale. 
We observe an upward ability trend as the model size increases, suggesting that the scaling law continues to hold. 
However, the ability variance among model families of the same size is smaller for the \textit{Following} task than for the \textit{Planning} task, indicating that the \textit{Following} ability is more stable and converged across model families.
The relatively uniform following ability across models indicates that, given a well-structured question, models of the same size can effectively leverage visual cues and contextual information to reach similar outcomes. 
This finding underscores the critical role of \textit{Planning} as a distinguishing factor in multi-step visual reasoning, particularly for problems that require more open-ended exploration strategies. 

\begin{findingbox}[title={Finding 3}]
\textit{Following} and \textit{Planning} abilities both contribute positively to the overall performance of the model.   
\end{findingbox}
\begin{figure}[h]
\centering
\includegraphics[width=0.9\linewidth]{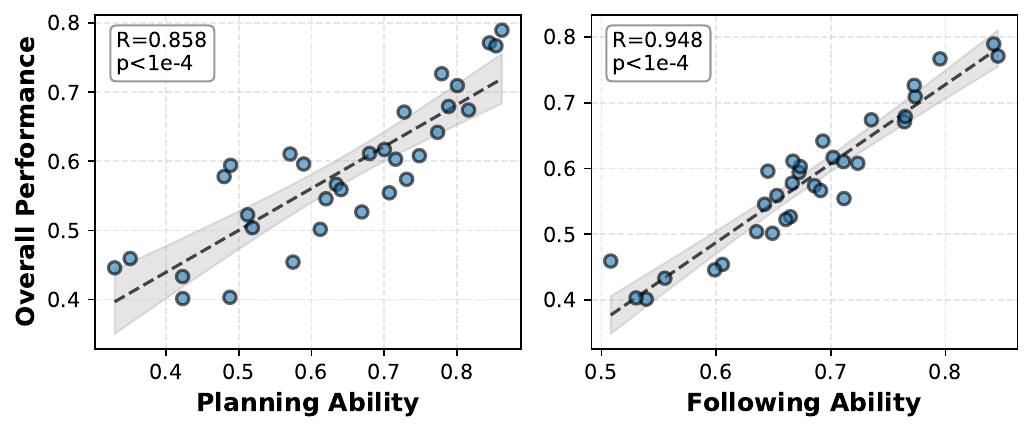}
\caption{Correlation between \textit{Following} and Overall performance (left), and between \textit{Planning} and Overall performance (right). \textbf{Both \textit{Following} and \textit{Planning} abilities are positively correlated with the overall performance of models.}}
\label{fig:correlation}
\end{figure}

We examine how a model’s \textit{Planning} and \textit{Following} abilities relate to its overall performance. 
Here, overall performance refers to model accuracy in directly answering the final question of each sample, which captures its end-to-end task competence.
As depicted in Figure~\ref{fig:correlation}, both the \textit{Following}-overall and \textit{Planning}-overall correlations are markedly strong, indicating that improvements in either dimension tend to yield gains in overall performance.
Quantitatively, the Pearson correlation coefficient between \textit{Following} ability and overall performance is $0.948$, while that between \textit{Planning} ability and overall performance is $0.858$. 
These results indicate that \textit{Following} ability remains a primary determinant of model end-to-end capability. 
However, the larger variance in \textit{Planning} ability across VLMs suggests that current models exhibit more pronounced disparities in their capacity for strategic reasoning. \looseness-1

\begin{findingbox}[title={Finding 4}]
Smaller models are better at \textit{Following} than \textit{Planning} while larger models have more balanced performance. 
\end{findingbox}

\begin{figure}[t]
\centering
\includegraphics[width=0.6\linewidth]{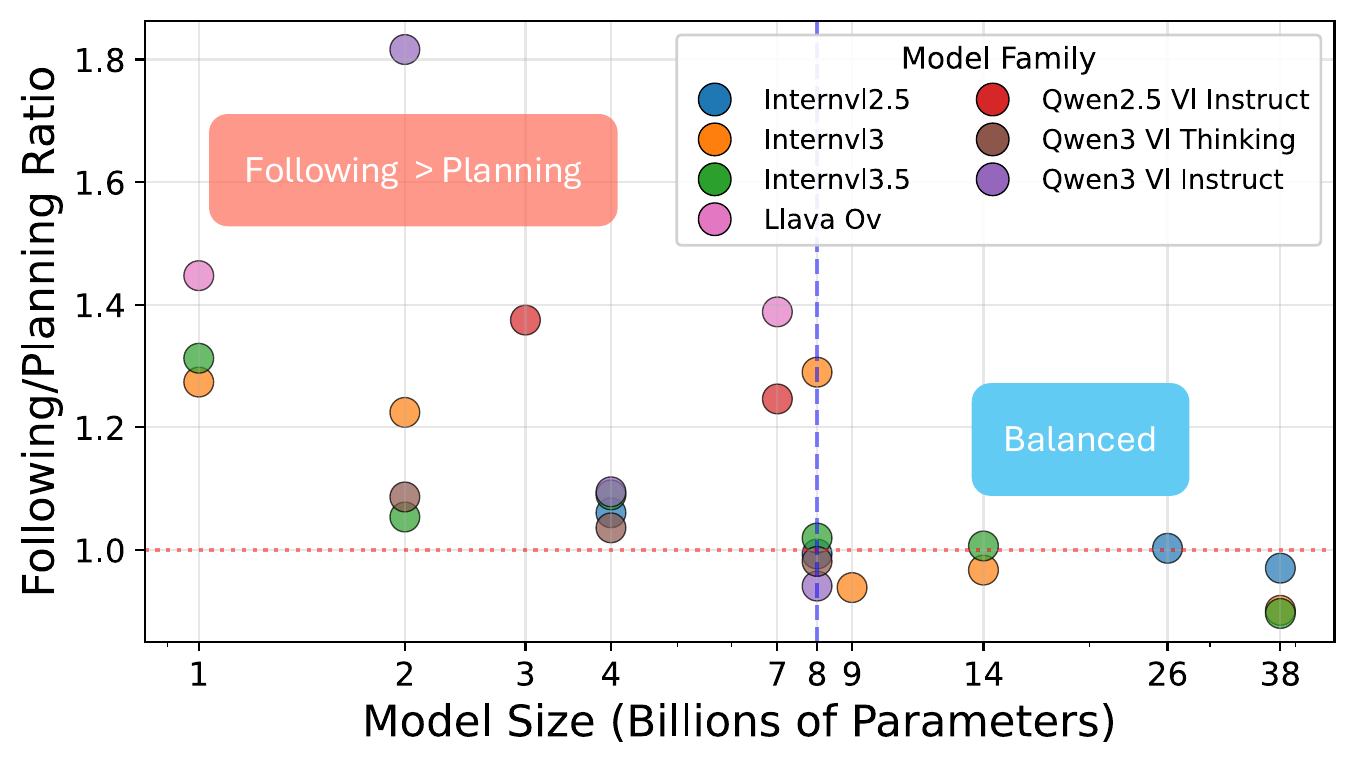}
\caption{\textbf{The ratio between \textit{Following} and \textit{Planning} ability at different model sizes. }
The abilities of \textit{Planning} and \textit{Following} are more balanced when the ratios are closer to $1.0$. 
\textbf{Smaller models are better at \textit{Following} than \textit{Planning} while larger models have more balanced ability.}}
\label{fig:balance}
\vspace{-4mm}
\end{figure}

Figure~\ref{fig:balance} presents the ratio of \textit{Following} to \textit{Planning} ability as a function of model size. 
The results indicate that smaller models exhibit a pronounced imbalance, with substantially greater proficiency in \textit{Following} compared to \textit{Planning}. 
As model size increases, however, this ratio approaches unity and stabilizes, reflecting that larger models achieve a more balanced performance across both reasoning dimensions. 
This trend suggests that scaling model capacity enhances not only overall performance but also enables models to more evenly develop planning-oriented reasoning capabilities alongside \textit{Following} skills.

\begin{findingbox}[title={Finding 5}]
VLMs are better at recovering from failed {planning} steps than from failed {following} steps.   
\end{findingbox}

We assess models' capability to recover from failure, defined as their ability to reach the correct final answer despite making at least one incorrect intermediate step.
This ability is crucial in real-world settings, where models often take suboptimal reasoning paths.
Table~\ref{tab:recover} reports failure-recovery performance for models with more than 10B parameters. 
The results indicate that all models exhibit some ability to recover, with recovery from failed {planning} generally exceeding recovery from failed {following}. 
This suggests that models are relatively more robust to suboptimal plans.
Distracting questions provide fewer clues but do not necessarily derail the final answer. 
In contrast, errors in the answer space are more likely to propagate to an incorrect final answer.

Another notable observation is that recovery from the failed {following} process does not differ significantly between smaller open-source models and larger proprietary models. 
By contrast, recovery from failed {planning} is markedly stronger in larger proprietary models. 
This indicates that larger proprietary models are less susceptible to suboptimal plans and to distraction from uninformative cues along the reasoning chain. 

\begin{wraptable}{r}{0.6\textwidth}
    \centering
    \vspace{-35pt}
    \caption{\textbf{VLMs' ability to recover from exploration failures}. The percentages of each model's responses that contain at least one wrong intermediate step but still arrive at correct final answers.}
    \label{tab:recover}
    \resizebox{\linewidth}{!}{\begin{tabular}{l|c|c}
    \toprule
    \textbf{Model} & From Failed {Planning} & From Failed {Following} \\
    \midrule
    \multicolumn{3}{c}{VLMs: $>\!$10B} \\
    \midrule
    InternVL3-14B & 65.9 & 55.9 \\
    InternVL3.5-14B & 65.5 & 52.6 \\
    InternVL2.5-26B & 66.9 & 62.1 \\
    InternVL2.5-38B & 69.2 & 63.2 \\
    InternVL3-38B & 65.1 & 52.9 \\
    InternVL3.5-38B & 65.4 & 45.6 \\
    \midrule
    \multicolumn{3}{c}{VLMs: Proprietary} \\
    \midrule
    GPT-4o & 78.9 & 56.5 \\
    GPT-5 & 84.3 & 53.6 \\
    o1 & 79.0 & 59.5 \\
    o3 & 80.6 & 57.3 \\
    Gemini 2.0 Flash & 69.5 & 59.4 \\
    Gemini 2.5 Flash & 72.7 & 51.8 \\
    \bottomrule
    \end{tabular}}
    \vspace{-4mm}
\end{wraptable}

\subsection{Additional Results}
Complementary results and experimental details are provided in the Appendix.
We establish a blindfold baseline in Appendix~\ref{sec:blindfold} to verify the visual-centric nature of our benchmark.
To accommodate future model advancements, we investigate increased distractor complexity in Appendix~\ref{sec:planning_8opts}, demonstrating the benchmark's potential for extensibility.
Finally, Appendix~\ref{appendix:cases_study} presents fine-grained case studies to highlight the diagnostic capabilities of our benchmark.











\section{Conclusion}

We introduced \ours, a benchmark that evaluates multi-step exploratory visual reasoning through a structured Chain-of-Questions (CoQ) formulation. By disentangling reasoning into \textit{Planning} and \textit{Following}, \ours provides fine-grained insights that are hidden in end-to-end evaluation. Experiments on different VLMs reveal consistent scaling trends, strong correlations between intermediate abilities and final performance, and a notable imbalance in smaller models, favoring \textit{Following} over \textit{Planning}. Our findings highlight exploratory reasoning, especially effective planning, as a key challenge for future VLMs and establish \ours as a foundation for advancing deliberate, stepwise visual reasoning. 

\bibliographystyle{splncs04}
\bibliography{main}

@String(CVPR= {IEEE Conf. Comput. Vis. Pattern Recog.})

@String(ICCV= {Int. Conf. Comput. Vis.})

@String(CVPR  = {CVPR})

@String(ICCV  = {ICCV})

@misc{liang2025colorbenchvlmsunderstandcolorful,
      title={ColorBench: Can VLMs See and Understand the Colorful World? A Comprehensive Benchmark for Color Perception, Reasoning, and Robustness}, 
      author={Yijun Liang and Ming Li and Chenrui Fan and Ziyue Li and Dang Nguyen and Kwesi Cobbina and Shweta Bhardwaj and Jiuhai Chen and Fuxiao Liu and Tianyi Zhou},
      year={2025},
      eprint={2504.10514},
      archivePrefix={arXiv},
      primaryClass={cs.CV},
      url={https://arxiv.org/abs/2504.10514}, 
}

@misc{openai2024openaio1card,
      title={OpenAI o1 System Card}, 
      author={OpenAI and : and Aaron Jaech and Adam Kalai and Adam Lerer and Adam Richardson and Ahmed El-Kishky and Aiden Low and Alec Helyar and et al.},
      year={2024},
      eprint={2412.16720},
      archivePrefix={arXiv},
      primaryClass={cs.AI},
      url={https://arxiv.org/abs/2412.16720}, 
}

@misc{li2025caughtcheatingmllmgoodcheating,
      title={CaughtCheating: Is Your MLLM a Good Cheating Detective? Exploring the Boundary of Visual Perception and Reasoning}, 
      author={Ming Li and Chenguang Wang and Yijun Liang and Xiyao Wang and Yuhang Zhou and Xiyang Wu and Yuqing Zhang and Ruiyi Zhang and Tianyi Zhou},
      year={2025},
      eprint={2507.00045},
      archivePrefix={arXiv},
      primaryClass={cs.CV},
      url={https://arxiv.org/abs/2507.00045}, 
}

@misc{chen2025visrbenchempiricalstudyvisual,
      title={VisR-Bench: An Empirical Study on Visual Retrieval-Augmented Generation for Multilingual Long Document Understanding}, 
      author={Jian Chen and Ming Li and Jihyung Kil and Chenguang Wang and Tong Yu and Ryan Rossi and Tianyi Zhou and Changyou Chen and Ruiyi Zhang},
      year={2025},
      eprint={2508.07493},
      archivePrefix={arXiv},
      primaryClass={cs.CV},
      url={https://arxiv.org/abs/2508.07493}, 
}

@misc{li2025visualtextgroundingmultimodal,
      title={Towards Visual Text Grounding of Multimodal Large Language Model}, 
      author={Ming Li and Ruiyi Zhang and Jian Chen and Chenguang Wang and Jiuxiang Gu and Yufan Zhou and Franck Dernoncourt and Wanrong Zhu and Tianyi Zhou and Tong Sun},
      year={2025},
      eprint={2504.04974},
      archivePrefix={arXiv},
      primaryClass={cs.CV},
      url={https://arxiv.org/abs/2504.04974}, 
}

@misc{roberts2025zerobenchimpossiblevisualbenchmark,
      title={ZeroBench: An Impossible Visual Benchmark for Contemporary Large Multimodal Models}, 
      author={Jonathan Roberts and Mohammad Reza Taesiri and Ansh Sharma and Akash Gupta and Samuel Roberts and Ioana Croitoru and Simion-Vlad Bogolin and Jialu Tang and Florian Langer and Vyas Raina and Vatsal Raina and Hanyi Xiong and Vishaal Udandarao and Jingyi Lu and Shiyang Chen and Sam Purkis and Tianshuo Yan and Wenye Lin and Gyungin Shin and Qiaochu Yang and Anh Totti Nguyen and David I. Atkinson and Aaditya Baranwal and Alexandru Coca and Mikah Dang and Sebastian Dziadzio and Jakob D. Kunz and Kaiqu Liang and Alexander Lo and Brian Pulfer and Steven Walton and Charig Yang and Kai Han and Samuel Albanie},
      year={2025},
      eprint={2502.09696},
      archivePrefix={arXiv},
      primaryClass={cs.CV},
      url={https://arxiv.org/abs/2502.09696}, 
}

@misc{shen2025vlmr1stablegeneralizabler1style,
      title={VLM-R1: A Stable and Generalizable R1-style Large Vision-Language Model}, 
      author={Haozhan Shen and Peng Liu and Jingcheng Li and Chunxin Fang and Yibo Ma and Jiajia Liao and Qiaoli Shen and Zilun Zhang and Kangjia Zhao and Qianqian Zhang and Ruochen Xu and Tiancheng Zhao},
      year={2025},
      eprint={2504.07615},
      archivePrefix={arXiv},
      primaryClass={cs.CV},
      url={https://arxiv.org/abs/2504.07615}, 
}

@misc{tan2025reasonrft,
      title={Reason-RFT: Reinforcement Fine-Tuning for Visual Reasoning of Vision Language Models}, 
      author={Huajie Tan and Yuheng Ji and Xiaoshuai Hao and Xiansheng Chen and Pengwei Wang and Zhongyuan Wang and Shanghang Zhang},
      year={2025},
      eprint={2503.20752},
      archivePrefix={arXiv},
      primaryClass={cs.CV},
      url={https://arxiv.org/abs/2503.20752}, 
}

@inproceedings{thawakar-etal-2025-llamav,
    title = "{L}lama{V}-o1: Rethinking Step-by-step Visual Reasoning in {LLM}s",
    author = "Thawakar, Omkar  and
      Dissanayake, Dinura  and
      More, Ketan Pravin  and
      Thawkar, Ritesh  and
      Heakl, Ahmed  and
      Ahsan, Noor  and
      Li, Yuhao  and
      Zumri, Ilmuz Zaman Mohammed  and
      Lahoud, Jean  and
      Anwer, Rao Muhammad  and
      Cholakkal, Hisham  and
      Laptev, Ivan  and
      Shah, Mubarak  and
      Khan, Fahad Shahbaz  and
      Khan, Salman",
    editor = "Che, Wanxiang  and
      Nabende, Joyce  and
      Shutova, Ekaterina  and
      Pilehvar, Mohammad Taher",
    booktitle = "Findings of the Association for Computational Linguistics: ACL 2025",
    month = jul,
    year = "2025",
    address = "Vienna, Austria",
    publisher = "Association for Computational Linguistics",
    url = "https://aclanthology.org/2025.findings-acl.1247/",
    doi = "10.18653/v1/2025.findings-acl.1247",
    pages = "24290--24315",
    ISBN = "979-8-89176-256-5"
}

@misc{xu2025visulogicbenchmarkevaluatingvisual,
      title={VisuLogic: A Benchmark for Evaluating Visual Reasoning in Multi-modal Large Language Models}, 
      author={Weiye Xu and Jiahao Wang and Weiyun Wang and Zhe Chen and Wengang Zhou and Aijun Yang and Lewei Lu and Houqiang Li and Xiaohua Wang and Xizhou Zhu and Wenhai Wang and Jifeng Dai and Jinguo Zhu},
      year={2025},
      eprint={2504.15279},
      archivePrefix={arXiv},
      primaryClass={cs.CV},
      url={https://arxiv.org/abs/2504.15279}, 
}

@InProceedings{Dong_2025_CVPR,
    author    = {Dong, Yuhao and Liu, Zuyan and Sun, Hai-Long and Yang, Jingkang and Hu, Winston and Rao, Yongming and Liu, Ziwei},
    title     = {Insight-V: Exploring Long-Chain Visual Reasoning with Multimodal Large Language Models},
    booktitle = {Proceedings of the IEEE/CVF Conference on Computer Vision and Pattern Recognition (CVPR)},
    month     = {June},
    year      = {2025},
    pages     = {9062-9072}
}

@misc{zhou2025r1zerosahamomentvisual,
      title={R1-Zero's "Aha Moment" in Visual Reasoning on a 2B Non-SFT Model}, 
      author={Hengguang Zhou and Xirui Li and Ruochen Wang and Minhao Cheng and Tianyi Zhou and Cho-Jui Hsieh},
      year={2025},
      eprint={2503.05132},
      archivePrefix={arXiv},
      primaryClass={cs.AI},
      url={https://arxiv.org/abs/2503.05132}, 
}

@misc{deepseekai2025deepseekr1incentivizingreasoningcapability,
      title={DeepSeek-R1: Incentivizing Reasoning Capability in LLMs via Reinforcement Learning}, 
      author={DeepSeek-AI and Daya Guo and Dejian Yang and Haowei Zhang and Junxiao Song and Ruoyu Zhang and Runxin Xu and Qihao Zhu and et al.},
      year={2025},
      eprint={2501.12948},
      archivePrefix={arXiv},
      primaryClass={cs.CL},
      url={https://arxiv.org/abs/2501.12948}, 
}

@inproceedings{satar-etal-2025-seeing,
    title = "Seeing Culture: A Benchmark for Visual Reasoning and Grounding",
    author = "Satar, Burak  and
      Ma, Zhixin  and
      Irawan, Patrick Amadeus  and
      Mulyawan, Wilfried Ariel  and
      Jiang, Jing  and
      Lim, Ee-Peng  and
      Ngo, Chong-Wah",
    editor = "Christodoulopoulos, Christos  and
      Chakraborty, Tanmoy  and
      Rose, Carolyn  and
      Peng, Violet",
    booktitle = "Proceedings of the 2025 Conference on Empirical Methods in Natural Language Processing",
    month = nov,
    year = "2025",
    address = "Suzhou, China",
    publisher = "Association for Computational Linguistics",
    url = "https://aclanthology.org/2025.emnlp-main.1131/",
    doi = "10.18653/v1/2025.emnlp-main.1131",
    pages = "22238--22254",
    ISBN = "979-8-89176-332-6"
}

@misc{cheng2025evaluatingmllmsmultimodalmultiimage,
      title={Evaluating MLLMs with Multimodal Multi-image Reasoning Benchmark}, 
      author={Ziming Cheng and Binrui Xu and Lisheng Gong and Zuhe Song and Tianshuo Zhou and Shiqi Zhong and Siyu Ren and Mingxiang Chen and Xiangchao Meng and Yuxin Zhang and Yanlin Li and Lei Ren and Wei Chen and Zhiyuan Huang and Mingjie Zhan and Xiaojie Wang and Fangxiang Feng},
      year={2025},
      eprint={2506.04280},
      archivePrefix={arXiv},
      primaryClass={cs.CV},
      url={https://arxiv.org/abs/2506.04280}, 
}

@inproceedings{pandya-etal-2025-ntsebench,
    title = "{NTSEBENCH}: Cognitive Reasoning Benchmark for Vision Language Models",
    author = "Pandya, Pranshu  and
      Gupta, Vatsal  and
      Talwarr, Agney S  and
      Kataria, Tushar  and
      Roth, Dan  and
      Gupta, Vivek",
    editor = "Chiruzzo, Luis  and
      Ritter, Alan  and
      Wang, Lu",
    booktitle = "Findings of the Association for Computational Linguistics: NAACL 2025",
    month = apr,
    year = "2025",
    address = "Albuquerque, New Mexico",
    publisher = "Association for Computational Linguistics",
    url = "https://aclanthology.org/2025.findings-naacl.204/",
    doi = "10.18653/v1/2025.findings-naacl.204",
    pages = "3680--3708",
    ISBN = "979-8-89176-195-7"
}

@misc{ren2025vgrpbenchvisualgridreasoning,
      title={VGRP-Bench: Visual Grid Reasoning Puzzle Benchmark for Large Vision-Language Models}, 
      author={Yufan Ren and Konstantinos Tertikas and Shalini Maiti and Junlin Han and Tong Zhang and Sabine Süsstrunk and Filippos Kokkinos},
      year={2025},
      eprint={2503.23064},
      archivePrefix={arXiv},
      primaryClass={cs.CV},
      url={https://arxiv.org/abs/2503.23064}, 
}

@misc{cai2025comparebenchbenchmarkvisualcomparison,
      title={CompareBench: A Benchmark for Visual Comparison Reasoning in Vision-Language Models}, 
      author={Jie Cai and Kangning Yang and Lan Fu and Jiaming Ding and Jinlong Li and Huiming Sun and Daitao Xing and Jinglin Shen and Zibo Meng},
      year={2025},
      eprint={2509.22737},
      archivePrefix={arXiv},
      primaryClass={cs.CV},
      url={https://arxiv.org/abs/2509.22737}, 
}

@misc{chen2025unveilingchainstepreasoning,
      title={Unveiling Chain of Step Reasoning for Vision-Language Models with Fine-grained Rewards}, 
      author={Honghao Chen and Xingzhou Lou and Xiaokun Feng and Kaiqi Huang and Xinlong Wang},
      year={2025},
      eprint={2509.19003},
      archivePrefix={arXiv},
      primaryClass={cs.CV},
      url={https://arxiv.org/abs/2509.19003}, 
}

@misc{openai2024gpt4ocard,
      title={GPT-4o System Card}, 
      author={OpenAI and Aaron Hurst and Adam Lerer and Adam P. Goucher and Adam Perelman and Aditya Ramesh and Aidan Clark and AJ Ostrow and Akila Welihinda and Alan Hayes and Alec Radford and etc.},
      year={2024},
      eprint={2410.21276},
      archivePrefix={arXiv},
      primaryClass={cs.CL},
      url={https://arxiv.org/abs/2410.21276}, 
}

@article{jaech2024openai,
  title={Openai o1 system card},
  author={Jaech, Aaron and Kalai, Adam and Lerer, Adam and Richardson, Adam and El-Kishky, Ahmed and Low, Aiden and Helyar, Alec and Madry, Aleksander and Beutel, Alex and Carney, Alex and others},
  journal={arXiv preprint arXiv:2412.16720},
  year={2024}
}

@misc{openai2025gpt5,
  title={GPT-5 System Card},
  author={OpenAI},
  year={2025},
  month={August},
  url={https://cdn.openai.com/gpt-5-system-card.pdf}
}

@techreport{openai2025o3o4mini,
  title       = {OpenAI o3 and o4-mini System Card},
  author      = {OpenAI},
  institution = {OpenAI},
  year        = {2025},
  month       = {April},
  url         = {https://cdn.openai.com/pdf/2221c875-02dc-4789-800b-e7758f3722c1/o3-and-o4-mini-system-card.pdf}
}

@misc{deepmind_gemini_flash,
  author       = {Google DeepMind},
  title        = {Gemini 2.0 Flash},
  year         = 2025,
  url          = {https://deepmind.google/technologies/gemini/flash/},
}

@article{comanici2025gemini,
  title={Gemini 2.5: Pushing the frontier with advanced reasoning, multimodality, long context, and next generation agentic capabilities},
  author={Comanici, Gheorghe and Bieber, Eric and Schaekermann, Mike and Pasupat, Ice and Sachdeva, Noveen and Dhillon, Inderjit and Blistein, Marcel and Ram, Ori and Zhang, Dan and Rosen, Evan and others},
  journal={arXiv preprint arXiv:2507.06261},
  year={2025}
}

@misc{li2024llavaonevisioneasyvisualtask,
      title={LLaVA-OneVision: Easy Visual Task Transfer}, 
      author={Bo Li and Yuanhan Zhang and Dong Guo and Renrui Zhang and Feng Li and Hao Zhang and Kaichen Zhang and Peiyuan Zhang and Yanwei Li and Ziwei Liu and Chunyuan Li},
      year={2024},
      eprint={2408.03326},
      archivePrefix={arXiv},
      primaryClass={cs.CV},
      url={https://arxiv.org/abs/2408.03326}, 
}

@inproceedings{chen2024internvl,
  title={Internvl: Scaling up vision foundation models and aligning for generic visual-linguistic tasks},
  author={Chen, Zhe and Wu, Jiannan and Wang, Wenhai and Su, Weijie and Chen, Guo and Xing, Sen and Zhong, Muyan and Zhang, Qinglong and Zhu, Xizhou and Lu, Lewei and others},
  booktitle={Proceedings of the IEEE/CVF Conference on Computer Vision and Pattern Recognition},
  pages={24185--24198},
  year={2024}
}

@article{chen2024expanding,
  title={Expanding performance boundaries of open-source multimodal models with model, data, and test-time scaling},
  author={Chen, Zhe and Wang, Weiyun and Cao, Yue and Liu, Yangzhou and Gao, Zhangwei and Cui, Erfei and Zhu, Jinguo and Ye, Shenglong and Tian, Hao and Liu, Zhaoyang and others},
  journal={arXiv preprint arXiv:2412.05271},
  year={2024}
}

@misc{bai2025qwen25vltechnicalreport,
      title={Qwen2.5-VL Technical Report}, 
      author={Shuai Bai and Keqin Chen and Xuejing Liu and Jialin Wang and Wenbin Ge and Sibo Song and Kai Dang and Peng Wang and Shijie Wang and Jun Tang and Humen Zhong and Yuanzhi Zhu and Mingkun Yang and Zhaohai Li and Jianqiang Wan and Pengfei Wang and Wei Ding and Zheren Fu and Yiheng Xu and Jiabo Ye and Xi Zhang and Tianbao Xie and Zesen Cheng and Hang Zhang and Zhibo Yang and Haiyang Xu and Junyang Lin},
      year={2025},
      eprint={2502.13923},
      archivePrefix={arXiv},
      primaryClass={cs.CV},
      url={https://arxiv.org/abs/2502.13923}, 
}

@article{zhu2025internvl3,
  title={Internvl3: Exploring advanced training and test-time recipes for open-source multimodal models},
  author={Zhu, Jinguo and Wang, Weiyun and Chen, Zhe and Liu, Zhaoyang and Ye, Shenglong and Gu, Lixin and Tian, Hao and Duan, Yuchen and Su, Weijie and Shao, Jie and others},
  journal={arXiv preprint arXiv:2504.10479},
  year={2025}
}

@article{wang2025internvl3,
  title={Internvl3. 5: Advancing open-source multimodal models in versatility, reasoning, and efficiency},
  author={Wang, Weiyun and Gao, Zhangwei and Gu, Lixin and Pu, Hengjun and Cui, Long and Wei, Xingguang and Liu, Zhaoyang and Jing, Linglin and Ye, Shenglong and Shao, Jie and others},
  journal={arXiv preprint arXiv:2508.18265},
  year={2025}
}

@article{yang2025qwen3,
  title={Qwen3 technical report},
  author={Yang, An and Li, Anfeng and Yang, Baosong and Zhang, Beichen and Hui, Binyuan and Zheng, Bo and Yu, Bowen and Gao, Chang and Huang, Chengen and Lv, Chenxu and others},
  journal={arXiv preprint arXiv:2505.09388},
  year={2025}
}

@article{chow2025physbench,
  title={Physbench: Benchmarking and enhancing vision-language models for physical world understanding},
  author={Chow, Wei and Mao, Jiageng and Li, Boyi and Seita, Daniel and Guizilini, Vitor and Wang, Yue},
  journal={arXiv preprint arXiv:2501.16411},
  year={2025}
}

@article{wang2024measuring,
  title={Measuring multimodal mathematical reasoning with math-vision dataset},
  author={Wang, Ke and Pan, Junting and Shi, Weikang and Lu, Zimu and Ren, Houxing and Zhou, Aojun and Zhan, Mingjie and Li, Hongsheng},
  journal={Advances in Neural Information Processing Systems},
  volume={37},
  pages={95095--95169},
  year={2024}
}

@article{tong2024cambrian,
  title={Cambrian-1: A fully open, vision-centric exploration of multimodal llms},
  author={Tong, Peter and Brown, Ellis and Wu, Penghao and Woo, Sanghyun and IYER, Adithya Jairam Vedagiri and Akula, Sai Charitha and Yang, Shusheng and Yang, Jihan and Middepogu, Manoj and Wang, Ziteng and others},
  journal={Advances in Neural Information Processing Systems},
  volume={37},
  pages={87310--87356},
  year={2024}
}

@inproceedings{mathew2022infographicvqa,
  title={Infographicvqa},
  author={Mathew, Minesh and Bagal, Viraj and Tito, Rub{\`e}n and Karatzas, Dimosthenis and Valveny, Ernest and Jawahar, CV},
  booktitle={Proceedings of the IEEE/CVF Winter Conference on Applications of Computer Vision},
  pages={1697--1706},
  year={2022}
}

@misc{marcelomoreno26_geoguessr_2024,
  author       = {Marcelo Moreno},
  title        = {geoguessr},
  year         = {2024},
  publisher    = {Hugging Face},
  url          = {https://huggingface.co/datasets/marcelomoreno26/geoguessr}
}

@misc{sd16_timeguessr_dataset,
  author       = {Saloni Dalal},
  title        = {timeguessr},
  year         = {2025},
  publisher    = {Hugging Face},
  url          = {https://huggingface.co/datasets/sd16/timeguessr}
}

@inproceedings{idealgpt_emnlp2023,
    title = "{I}deal{GPT}: Iteratively Decomposing Vision and Language Reasoning via Large Language Models",
    author = "You, Haoxuan  and
      Sun, Rui  and
      Wang, Zhecan  and
      Chen, Long  and
      Wang, Gengyu  and
      Ayyubi, Hammad  and
      Chang, Kai-Wei  and
      Chang, Shih-Fu",
    editor = "Bouamor, Houda  and
      Pino, Juan  and
      Bali, Kalika",
    booktitle = "Findings of the Association for Computational Linguistics: EMNLP 2023",
    month = dec,
    year = "2023",
    address = "Singapore",
    publisher = "Association for Computational Linguistics",
    url = "https://aclanthology.org/2023.findings-emnlp.755/",
    doi = "10.18653/v1/2023.findings-emnlp.755",
    pages = "11289--11303"
}

@InProceedings{llavacot_iccv2025,
    author    = {Xu, Guowei and Jin, Peng and Wu, Ziang and Li, Hao and Song, Yibing and Sun, Lichao and Yuan, Li},
    title     = {LLaVA-CoT: Let Vision Language Models Reason Step-by-Step},
    booktitle = {Proceedings of the IEEE/CVF International Conference on Computer Vision (ICCV)},
    month     = {October},
    year      = {2025},
    pages     = {2087-2098}
}

@inproceedings{
limitVLMs_neurips24,
title={Understanding the Limits of Vision Language Models Through the Lens of the Binding Problem},
author={Declan Iain Campbell and Sunayana Rane and Tyler Giallanza and C. Nicol{\`o} De Sabbata and Kia Ghods and Amogh Joshi and Alexander Ku and Steven M Frankland and Thomas L. Griffiths and Jonathan D. Cohen and Taylor Whittington Webb},
booktitle={The Thirty-eighth Annual Conference on Neural Information Processing Systems},
year={2024},
url={https://openreview.net/forum?id=Q5RYn6jagC}
}

@InProceedings{capture_iccv2025,
    author    = {Pothiraj, Atin and Stengel-Eskin, Elias and Cho, Jaemin and Bansal, Mohit},
    title     = {CAPTURE: Evaluating Spatial Reasoning in Vision Language Models via Occluded Object Counting},
    booktitle = {Proceedings of the IEEE/CVF International Conference on Computer Vision (ICCV)},
    month     = {October},
    year      = {2025},
    pages     = {8001-8010}
}

@inproceedings{
multistep_neurips25,
title={Multi-step Visual Reasoning with Visual Tokens Scaling and Verification},
author={Tianyi Bai and Zengjie Hu and Fupeng Sun and Qiu Jiantao and Yizhen Jiang and Guangxin He and Bohan Zeng and Conghui He and Binhang Yuan and Wentao Zhang},
booktitle={The Thirty-ninth Annual Conference on Neural Information Processing Systems},
year={2025},
url={https://openreview.net/forum?id=y60FhgO07j}
}

@misc{liu2024imagebasedgeolocationusinglarge,
      title={Image-Based Geolocation Using Large Vision-Language Models}, 
      author={Yi Liu and Junchen Ding and Gelei Deng and Yuekang Li and Tianwei Zhang and Weisong Sun and Yaowen Zheng and Jingquan Ge and Yang Liu},
      year={2024},
      eprint={2408.09474},
      archivePrefix={arXiv},
      primaryClass={cs.CR},
      url={https://arxiv.org/abs/2408.09474}, 
}

@misc{wang2025multimodalchainofthoughtreasoningcomprehensive,
      title={Multimodal Chain-of-Thought Reasoning: A Comprehensive Survey}, 
      author={Yaoting Wang and Shengqiong Wu and Yuecheng Zhang and Shuicheng Yan and Ziwei Liu and Jiebo Luo and Hao Fei},
      year={2025},
      eprint={2503.12605},
      archivePrefix={arXiv},
      primaryClass={cs.CV},
      url={https://arxiv.org/abs/2503.12605}, 
}

@misc{li2025perceptionreasonthinkplan,
      title={Perception, Reason, Think, and Plan: A Survey on Large Multimodal Reasoning Models}, 
      author={Yunxin Li and Zhenyu Liu and Zitao Li and Xuanyu Zhang and Zhenran Xu and Xinyu Chen and Haoyuan Shi and Shenyuan Jiang and Xintong Wang and Jifang Wang and Shouzheng Huang and Xinping Zhao and Borui Jiang and Lanqing Hong and Longyue Wang and Zhuotao Tian and Baoxing Huai and Wenhan Luo and Weihua Luo and Zheng Zhang and Baotian Hu and Min Zhang},
      year={2025},
      eprint={2505.04921},
      archivePrefix={arXiv},
      primaryClass={cs.CV},
      url={https://arxiv.org/abs/2505.04921}, 
}

@misc{lu2024mathvistaevaluatingmathematicalreasoning,
      title={MathVista: Evaluating Mathematical Reasoning of Foundation Models in Visual Contexts}, 
      author={Pan Lu and Hritik Bansal and Tony Xia and Jiacheng Liu and Chunyuan Li and Hannaneh Hajishirzi and Hao Cheng and Kai-Wei Chang and Michel Galley and Jianfeng Gao},
      year={2024},
      eprint={2310.02255},
      archivePrefix={arXiv},
      primaryClass={cs.CV},
      url={https://arxiv.org/abs/2310.02255}, 
}

@misc{wang2025benchmarkingmultimodalmathematicalreasoning,
      title={Benchmarking Multimodal Mathematical Reasoning with Explicit Visual Dependency}, 
      author={Zhikai Wang and Jiashuo Sun and Wenqi Zhang and Zhiqiang Hu and Xin Li and Fan Wang and Deli Zhao},
      year={2025},
      eprint={2504.18589},
      archivePrefix={arXiv},
      primaryClass={cs.CV},
      url={https://arxiv.org/abs/2504.18589}, 
}

@misc{lyu2025jigsawpuzzlesseeingunderstandingreasoning,
      title={Jigsaw-Puzzles: From Seeing to Understanding to Reasoning in Vision-Language Models}, 
      author={Zesen Lyu and Dandan Zhang and Wei Ye and Fangdi Li and Zhihang Jiang and Yao Yang},
      year={2025},
      eprint={2505.20728},
      archivePrefix={arXiv},
      primaryClass={cs.AI},
      url={https://arxiv.org/abs/2505.20728}, 
}

@misc{song2025visualpuzzlesdecouplingmultimodalreasoning,
      title={VisualPuzzles: Decoupling Multimodal Reasoning Evaluation from Domain Knowledge}, 
      author={Yueqi Song and Tianyue Ou and Yibo Kong and Zecheng Li and Graham Neubig and Xiang Yue},
      year={2025},
      eprint={2504.10342},
      archivePrefix={arXiv},
      primaryClass={cs.CL},
      url={https://arxiv.org/abs/2504.10342}, 
}
\clearpage


\appendix
\section{Image Sources}
\label{appendix:data_sources}

We collect images for \ours from a combination of website sources and publicly available visual reasoning benchmarks. 
Specifically, our dataset incorporates images from PhysBench\citep{chow2025physbench}, MathVision\citep{wang2024measuring}, CV-Bench\citep{tong2024cambrian}, and InfographicQA\citep{mathew2022infographicvqa}. 
In addition, we include real-world images sourced from the online platforms GeoGuessr\citep{marcelomoreno26_geoguessr_2024} and TimeGuessr\citep{sd16_timeguessr_dataset}.
Together, these diverse sources ensure broad coverage across the 15 application scenarios represented in \ours.
\section{Detailed Taxonomy}
\label{appendix:taxonomy}

\paragraph{Deduction.} 
This category addresses rule-based and logical reasoning, requiring models to explore potential causal or relational patterns to reach consistent conclusions. 
Exploration involves hypothesizing, verifying, and generalizing implicit rules from limited observations.
\textit{\textbf{Flowchart Deduction}} assesses whether models can reason through multi-step logical structures by tracing dependencies in flow diagrams. 
\textit{\textbf{Pattern Deduction}} examines the ability to identify mathematical or visual regularities and extrapolate unseen elements. 
\textit{\textbf{Property Deduction}} tests the capacity to infer physical or material properties based on observed patterns, and 
\textit{\textbf{Relationship Deduction}} measures understanding of inter-object relationships and causal interactions in structured visual data. 
Together, these tasks capture the model’s exploratory capacity to discover and apply underlying rules within multi-step reasoning processes.

\paragraph{Guessing.} 
This category examines reasoning under uncertainty, requiring models to explore incomplete or ambiguous evidence to infer missing information. 
Exploration is reflected in the model’s ability to generate and evaluate multiple hypotheses before settling on the most plausible one.
\textit{\textbf{Intention Guessing}} measures a model’s ability to infer human or object intentions from dynamic visual cues. 
\textit{\textbf{Location Guessing}} requires identifying plausible locations based on contextual evidence, while 
\textit{\textbf{Responsibility Guessing}} examines causal reasoning in scenarios such as traffic incidents or shared actions. 
\textit{\textbf{Time Guessing}} tests temporal inference based on environmental cues, and 
\textit{\textbf{Topic Guessing}} evaluates the ability to infer abstract or thematic concepts from multimodal contexts. 
These tasks together assess exploratory reasoning through uncertainty resolution and context-driven hypothesis testing.

\paragraph{Navigation.} 
This category highlights spatial and sequential reasoning, requiring models to explore action sequences, spatial configurations, and reasoning trajectories. 
Exploration is reflected in the model’s ability to plan, simulate, and adjust its reasoning path while preserving overall coherence.
\textit{\textbf{Map Navigation}} tests spatial reasoning and route planning, while 
\textit{\textbf{GUI Navigation}} evaluates the model’s ability to reason through interface layouts and procedural tasks. 
\textit{\textbf{Traffic Navigation}} focuses on understanding dynamic interactions under traffic scenarios and rules. 
\textit{\textbf{Trend Navigation}} examines the ability to interpret sequential or time-evolving visual patterns. 
Together, these tasks assess exploratory reasoning as models must simulate and adjust multi-step plans in dynamic spatial or temporal settings.

\paragraph{Retrieval.} 
This category emphasizes global evidence integration, prompting the model to explore multiple visual or textual components. 
Exploration emerges through the model’s capacity to locate, select, and combine dispersed cues into a coherent reasoning outcome.
\textit{\textbf{Counting Retrieval}} tests whether models can locate and aggregate quantitative information across visual regions, while 
\textit{\textbf{Word Puzzle Retrieval}} evaluates the ability to identify, interpret, and reason over embedded textual elements. 
These tasks together measure exploratory reasoning through multi-step evidence gathering and integration, assessing how effectively models combine local details into coherent global understanding.

Together, these four categories and fifteen sub-tasks form a comprehensive taxonomy for evaluating exploratory reasoning. 
They capture complementary aspects of how VLMs hypothesize, search, plan, and integrate knowledge across diverse multimodal reasoning scenarios.

\section{Data Curation for \textit{Planning}}
\label{appendix:curation_planning}

\subsection{Generation Process}
\label{appendix:planning_curation_process}

\begin{figure}[t]
\centering
\includegraphics[width=0.65\linewidth]{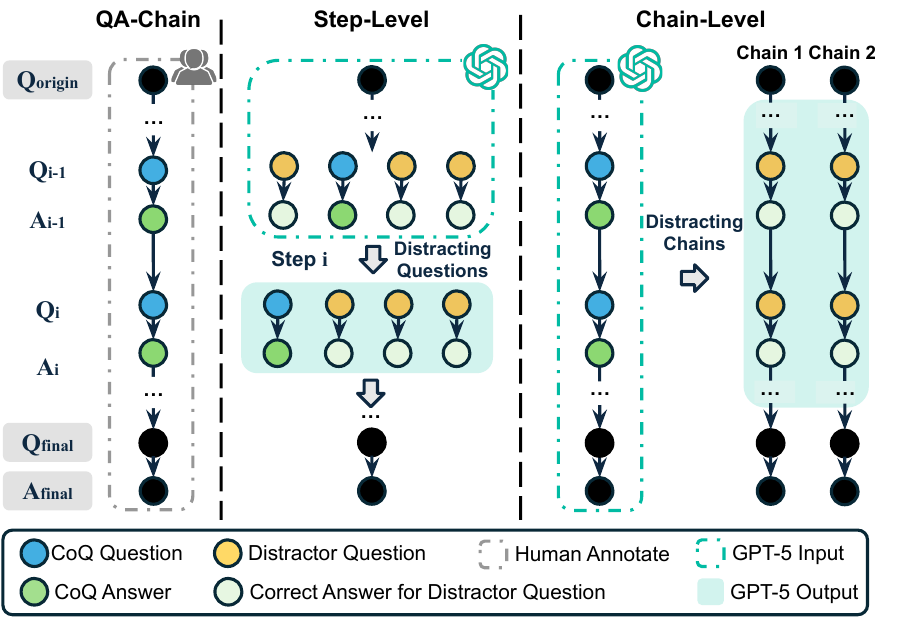}
\caption{\textbf{Generation Pipeline for \textit{Planning}}. For each QA chain, GPT-5 generates candidate distractors using two complementary strategies: (1) \textit{step-level}, which introduces diverse distracting questions at each reasoning step, and (2) \textit{chain-level}, which constructs full distracting reasoning chains. 
For each sample, we select $1$ most challenging distracting chain and $2$ diverse distracting questions per step to form the final distractor set.}
\label{fig:data_curation_planning}
\end{figure}

To evaluate VLMs’ exploratory abilities within the question space, we design a \textit{Planning} task in which models must navigate reasoning chains while confronted with multiple plausible yet distracting questions. 
This setup probes whether a model can select the most informative sub-questions amid competing alternatives.

The dataset is constructed using a two-stage pipeline composed of \textbf{Distractor Construction} and \textbf{Question Integration}. 
The pipeline uses powerful LLMs for generation, filtering, and refinement, resulting in question sets that are semantically rich, diverse, and challenging for VLMs.

\paragraph{Distractor Construction.}
This stage focuses on producing alternative questions that appear contextually relevant but deviate from the intended reasoning path.
We design two complementary strategies to capture both local and global variability within reasoning chains.
The first strategy, \textbf{\textit{step-level distractor generation}}, focuses on each reasoning step individually.
At every step, the LLM is prompted with the preceding context, including ground-truth questions and previously generated distractors, to propose new step-specific questions that remain plausible yet lead to divergent reasoning directions.
This encourages local diversity and tests a model's ability to identify the correct continuation among multiple confounding options.
The second strategy, \textbf{\textit{chain-level distractor generation}}, operates at a global level.
Given an entire ground-truth QA chain, the LLM generates multiple challenger chains that are logically coherent within themselves but deviate from the original reasoning trajectory.
These distractors evaluate whether a model can differentiate between globally consistent yet semantically incorrect reasoning paths.
To ensure higher-quality distractors, we employ GPT-5 for all question generation.

\paragraph{Question Integration.}
After generating both step-level and chain-level distractors, we integrate them into a unified Planning dataset through a refinement and selection process.
First, we use a strong open-sourced LLM (Qwen3-VL-32B-Instruct) to perform inference over a mixture of chain-level distractors and ground-truth questions, and automatically identify the chain that causes the highest confusion or reasoning inconsistency for each sample.
This model-guided evaluation highlights the most challenging and coherent distractor chains, which are then retained for integration.
Next, GPT-5 is prompted to select the most diverse and difficult step-level distractors that differ from the retained chains while preserving contextual plausibility.
These selected distractors are then merged with the original ground-truth questions to form the final multi-choice Planning dataset, where each reasoning step includes both correct and misleading alternatives.

Through this two-stage generation and integration pipeline, the \textit{Planning} task systematically probes VLMs' ability to explore and reason within complex question spaces.
It reveals whether models can plan consistent reasoning trajectories when confronted with subtle yet semantically misleading alternatives.

\subsection{Generation Prompt}
\label{appendix:planning_curation_prompt}

To support high-quality reasoning across both \textit{step-level} and \textit{chain-level} distractor generation, we design two structured system prompts that guide the closed-source GPT-5\citep{openai2025gpt5} model in producing challenging and misleading multi-step reasoning traces.
The prompts require the model to base every generated question on explicit visual evidence and to articulate a coherent progression of reasoning across steps, ensuring that each distractor remains contextually grounded while deviating from the intended reasoning path.
To accommodate different forms of reasoning complexity, the prompts specify detailed instructions for paraphrasing gold questions, generating plausible yet strategically irrelevant alternatives, and preserving internal coherence within each distractor chain.
Additionally, the prompts introduce explicit guidelines for separating task-relevant cues from unhelpful or misleading ones, preventing the model from inadvertently producing reasoning that supports the correct answer.

The prompts are further enhanced with internal checklists, stylistic constraints, and strict prohibitions against helpful reasoning patterns, enabling robust and controlled distractor generation.
By enforcing consistent output formats, grounding each question in the visual scene, and providing precise instructions for multi-step dependencies, these prompts facilitate the production of reliable paraphrased questions, step-level distractors, and globally coherent distracting chains.
This design supports interpretable and reproducible distractor construction, ensuring that the generated alternatives effectively challenge the reasoning capabilities of modern VLMs.

\begin{figure*}[t]
\begin{tcolorbox}[colback=gray!5, colframe=black!45,
    fonttitle=\bfseries, title={System Prompt for Step-Level Generation (Part 1)}]
\small
\lstset{breaklines=true, basicstyle=\footnotesize\ttfamily}

\begin{lstlisting}
You will be given:
1. A final question that must be answered based on an image.
2. (Optional) Intermediate questions and confusing questions for earlier steps.
   - If this is step 1, you will only receive the final question.
   - If this is step N > 1, you will receive all intermediate and confusing questions
     from steps 1 to N-1 as context.

Your task:
For the current reasoning step, generate 5 confusing or misleading questions that:
- Are visually and contextually relevant to the scene.
- Sound appropriate regardless of previous intermediate choices.
- Are NOT logically useful for solving the final answer.
- Should confuse the model's step planning.
- If the final question contains a condition ("If the input is xxx"),
  repeat that condition in the confusing questions for the first step.

In other words:
The confusing questions must appear contextually coherent but strategically irrelevant.
They should make an incorrect planning choice seem superficially reasonable.

For each unhelpful question, provide:
- A short factual answer.
- An unhelpfulness_score (1-3).
\end{lstlisting}

\end{tcolorbox}
\end{figure*}

\begin{figure*}[t]
\begin{tcolorbox}[colback=gray!5, colframe=black!45,
    fonttitle=\bfseries, title={System Prompt for Step-Level Generation (Part 2)}]
\small
\lstset{breaklines=true, basicstyle=\footnotesize\ttfamily}

\begin{lstlisting}
INTERNAL REASONING CHECKLIST (do not output):
1. Coherent with any prior intermediate steps.
2. Shares entities or context with the main task.
3. Does NOT provide causal, counting, spatial, or comparative reasoning.
4. Focuses on irrelevant attributes or secondary objects.
5. Appears plausible as a next-step question.

FORBIDDEN QUESTION TYPES:
- Overlap with any intermediate question's reasoning step.
- Provide causal, temporal, or numerical clues needed for the final answer.
- Help the model disambiguate the correct reasoning step.

Prefer "side-path" confusion questions:
- About background objects or irrelevant actions.
- About minor attributes (pose, color, expression, small details).
- About goals or comparisons unrelated to the final task.

OUTPUT RULES:
- Keep grounded, natural, image-based.
- No hallucinations.
- Maintain stylistic consistency.
- Return ONLY valid JSON:

{
  "question_N": [
    {"paraphrased_intermediate_question": "string", "answer": "string", "unhelpfulness_score": int},
    {"unhelpful_question": "string", "answer": "string", "unhelpfulness_score": int},
    {"unhelpful_question": "string", "answer": "string", "unhelpfulness_score": int},
    {"unhelpful_question": "string", "answer": "string", "unhelpfulness_score": int},
    ...
  ]
}
\end{lstlisting}

\end{tcolorbox}
\end{figure*}

\begin{figure*}[t]
\begin{tcolorbox}[colback=gray!5, colframe=black!45,
    fonttitle=\bfseries, title={System Prompt for Chain-Level Distractor Generation (Part 1)}]
\small
\lstset{breaklines=true, basicstyle=\small\ttfamily}

\begin{lstlisting}
You will be given:
1. An image.
2. A final question that must be answered based on the image.
3. Several intermediate questions that form the correct reasoning chain.
Your task:
Generate 2 confusing reasoning chains. For each gold step i, produce:
- A paraphrase of the gold intermediate question_i.
- One misleading question (and answer) for Chain 1.
- One misleading question (and answer) for Chain 2.
Each confusing question must:
- Be visually and contextually relevant to the image.
- Sound coherent and stylistically close to the intermediate/final question.
- Remain consistent across steps (a plausible but incorrect chain).
- Not be logically useful for solving the final question.
- Confuse the model's step planning by offering a tempting side-path.
- Repeat any conditional clause from the final question (e.g., "If the input is X")
  in the confusing questions for the first step.
INTERNAL REASONING CHECKLIST (do not output):
1. Shares scene context with the gold question.
2. Maintains a consistent distractor storyline.
3. Avoids reasoning patterns (causal, temporal, spatial, numerical, comparative)
   that lead to the correct answer.
4. Focuses on irrelevant or secondary attributes, actions, or objects.
\end{lstlisting}

\end{tcolorbox}
\end{figure*}

\begin{figure*}[t]
\begin{tcolorbox}[colback=gray!5, colframe=black!45,
    fonttitle=\bfseries, title={System Prompt for Chain-Level Distractor Generation (Part 2)}]
\small
\lstset{breaklines=true, basicstyle=\small\ttfamily}

\begin{lstlisting}
FORBIDDEN HELPFUL QUESTION TYPES:
- Repeating or clarifying any gold reasoning step.
- Providing causal, numerical, or spatial clues related to the final answer.
- Helping to disambiguate or verify the correct reasoning path.

PREFERRED QUESTION STYLE:
- "Side-path" questions about background details or secondary objects.
- Attributes and interactions unrelated to the final goal.
- Natural, grounded, and plausible but strategically irrelevant.

### OUTPUT RULES:
- Keep questions grounded and factual (no hallucinations).
- Avoid trivial perceptual details unless used as distractors.
- Avoid generating questions semantically identical to the original intermediate ones.
- PARAPHRASE the intermediate question to ensure stylistic consistency, return the original answer as the corresponding answer.
- Return ONLY valid JSON following the schema below.

### JSON OUTPUT SCHEMA:

{ "paraphrased_gt_questions": {"question_1": "string", "answer_1": string", ...},
"consistent_confusing_questions1": {"question_1": "string", "answer_1": string", ...},
"consistent_confusing_questions2": {"question_1": "string", "answer_1": string", ...}
}
\end{lstlisting}

\end{tcolorbox}
\end{figure*}

\clearpage
\section{Experiment Details}
\label{appendix:experiment}

\subsection{Implementation Details}
\label{appendix:implement}

To comprehensively evaluate the exploratory capabilities of VLMs, we assess a total of $32$ models spanning a broad range of state-of-the-art families, including both proprietary and open-source systems.
The evaluated models include GPT-4o\citep{openai2024gpt4ocard}, GPT-5\citep{openai2025gpt5}, O1\citep{jaech2024openai}, O3\citep{openai2025o3o4mini}, Gemini-2 Flash\citep{deepmind_gemini_flash}, Gemini-2.5 Flash\citep{comanici2025gemini}, LLaVA-OV\citep{li2024llavaonevisioneasyvisualtask}, InternVL2.5\citep{chen2024expanding}, InternVL3-Instruct\citep{zhu2025internvl3}, InternVL3.5-Instruct\citep{wang2025internvl3}, Qwen2.5-VL\citep{bai2025qwen25vltechnicalreport}, and Qwen3-VL-Instruct (IT) / Think\citep{yang2025qwen3}.
OpenAI and Gemini models are accessed through their official APIs, while all open-source models are locally deployed for evaluation.

This model suite spans a wide range of architectural designs and parameter scales, from $1$B to $38$B parameters, covering both multimodal proprietary systems and fully open-source implementations.
Such diversity enables a balanced and comprehensive assessment of exploratory reasoning behavior across different computational budgets and training paradigms.
To ensure a fair comparison, all open-source models are evaluated under a standardized experimental environment using a single NVIDIA A100 GPU (80GB), with identical decoding and temperature settings across all experiments.

\subsection{Evaluation Prompts}
\label{appendix:eval_prompts}

We provide $2$ prompts to assess model behavior under the \textit{Following} and \textit{Planning} tasks. 

For the \textit{Planning} task, the model is given the final question at the beginning of the full interaction. 
At each step, it receives the accumulated conversation history, the image, and a set of candidate intermediate questions consisting of both gold questions and distractors. 
The prompt requires the model to independently select the most helpful intermediate question from these alternatives and to output its choice in a precise, constrained format. 

For the \textit{Following} task, the model is provided with the conversation history, the image, and the corresponding question and answer candidates at each step. 
The prompt instructs the model to identify the correct answer for the given question, ensuring strict adherence to the human-designed reasoning chain.


\begin{figure}[h]
\begin{tcolorbox}[colback=gray!5, colframe=black!45,
    fonttitle=\bfseries, title={Evaluation Prompt for \textit{Planning} task}]
\small
\lstset{breaklines=true, basicstyle=\small\ttfamily}

\begin{lstlisting}
Choose the single most helpful intermediate question from the options below. On the FIRST line, output ONLY the option letter in parentheses (e.g., (A)). That first line must match exactly: ^\\([A-F]\\)$ After a blank line, you MAY add brief reasoning.
\end{lstlisting}

\end{tcolorbox}
\end{figure}

\begin{figure}[h]
\begin{tcolorbox}[colback=gray!5, colframe=black!45,
    fonttitle=\bfseries, title={Evaluation Prompt for \textit{Following} task}]
\small
\lstset{breaklines=true, basicstyle=\small\ttfamily}

\begin{lstlisting}
You need to choose the most likely answer from the following options, make sure the option letter is in the parentheses like (X).:
\end{lstlisting}

\end{tcolorbox}
\end{figure}

\section{Final Accuracy for VLMs on \ours}
\label{appendix:final_acc}

The accuracies on final questions under different settings (w/o CoQ, under the \textit{Planning} task, and under the \textit{Following} task) are shown in Table~\ref{tab:main_acc}.
Table~\ref{tab:main_acc} reports the final-question accuracies of all $32$ evaluated VLMs across these three evaluation modes.

Under the w/o CoQ setting ($\text{Acc}$), the model answers the final question directly without relying on any intermediate reasoning steps.
$\text{Acc}_{Plan}$ reflects the final-question accuracy after the model explores the question space by selecting the most informative intermediate question at each step in the presence of distractors.
$\text{Acc}_{Follow}$ measures the final-question accuracy after the model navigates the answer space by choosing the correct answer to each intermediate question along the ground-truth reasoning chain.

Together, these results provide a comprehensive view of how model performance changes when guided by CoQ-based reasoning, highlighting the role of structured exploration in improving visual reasoning.

\begin{table*}[t]
\centering
\caption{Final accuracy over $4$ categories and $32$ VLMs with and without CoQ. }
\label{tab:main_acc}
\resizebox{\linewidth}{!}{\begin{tabular}{l|ccc|ccc|ccc|ccc|ccc}
\toprule
\textbf{Model} & \multicolumn{3}{c|}{\textbf{Deduction}} & \multicolumn{3}{c|}{\textbf{Guessing}} & \multicolumn{3}{c|}{\textbf{Navigation}} & \multicolumn{3}{c|}{\textbf{Retrieval}} & \multicolumn{3}{c}{\textbf{Overall}} \\
\cmidrule(lr){2-4}\cmidrule(lr){5-7}\cmidrule(lr){8-10}\cmidrule(lr){11-13}\cmidrule(lr){14-16}
 & ${\text{Acc}}$ & ${\text{Acc}}_{Plan}$ & ${\text{Acc}}_{Follow}$ & ${\text{Acc}}$ & ${\text{Acc}}_{Plan}$ & ${\text{Acc}}_{Follow}$ & ${\text{Acc}}$ & ${\text{Acc}}_{Plan}$ & ${\text{Acc}}_{Follow}$ & ${\text{Acc}}$ & ${\text{Acc}}_{Plan}$ & ${\text{Acc}}_{Follow}$ & ${\text{Acc}}$ & ${\text{Acc}}_{Plan}$ & ${\text{Acc}}_{Follow}$ \\
\midrule
\multicolumn{16}{c}{VLMs: $<$7B} \\
\midrule
LLaVA-OV-1B & 26.4 & 27.5 & 30.8 & 50.4 & 49.6 & 47.9 & 58.2 & 55.1 & 59.2 & 48.8 & 48.8 & 48.8 & 45.9 & 45.2 & 46.7 \\
InternVL3-1B & 26.4 & 31.9 & 27.5 & 49.6 & 49.6 & 50.4 & 60.2 & 63.3 & 69.4 & 24.4 & 22.0 & 36.6 & 40.1 & 41.7 & 46.0 \\
InternVL3.5-1B & 29.7 & 30.8 & 37.4 & 43.0 & 34.7 & 48.8 & 59.2 & 56.1 & 77.6 & 41.5 & 29.3 & 41.5 & 43.3 & 37.7 & 51.3 \\
Qwen3-VL-2B-IT & 24.2 & 20.9 & 40.7 & \textbf{66.1} & 38.8 & 63.6 & 61.2 & 37.8 & 77.6 & 26.8 & 34.1 & 34.1 & 44.6 & 32.9 & 54.0 \\
Qwen3-VL-2B-Think & 33.0 & 42.9 & 40.7 & 44.6 & 51.2 & 51.2 & 39.8 & 55.1 & 62.2 & 43.9 & 53.7 & 41.5 & 40.3 & 50.7 & 48.9 \\
InternVL3-2B & 40.7 & 47.3 & 44.0 & 54.5 & 62.8 & 61.2 & 67.3 & 74.5 & 78.6 & 39.0 & 43.9 & 48.8 & 50.4 & 57.1 & 58.1 \\
InternVL3.5-2B & 36.3 & 40.7 & 40.7 & 42.1 & 54.5 & 53.7 & 52.0 & 80.6 & 70.4 & 51.2 & 43.9 & 31.7 & 45.4 & 54.9 & 49.1 \\
Qwen2.5-VL-3B & 47.5 & 40.7 & \textbf{55.4} & 62.6 & 58.7 & 60.2 & 71.4 & 73.5 & 79.6 & 56.1 & \textbf{58.5} & 53.7 & 59.4 & 57.8 & 62.2 \\
Qwen3-VL-4B-IT & 47.3 & 44.0 & 54.9 & 65.3 & 63.6 & \textbf{68.6} & 67.3 & 76.5 & 82.7 & \textbf{58.5} & \textbf{58.5} & 48.8 & \textbf{59.6} & 60.7 & \textbf{63.7} \\
Qwen3-VL-4B-Think & \textbf{51.6} & \textbf{54.9} & 48.4 & 57.0 & 56.2 & 61.2 & 63.3 & 69.4 & 71.4 & 46.3 & 53.7 & \textbf{63.4} & 54.6 & 58.5 & 61.1 \\
InternVL2.5-4B & 36.3 & 41.8 & 42.9 & 54.5 & 57.0 & 62.0 & 68.4 & 81.6 & 80.6 & 41.5 & \textbf{58.5} & 43.9 & 50.2 & 59.7 & 57.3 \\
InternVL3.5-4B & 45.1 & 49.5 & 48.4 & 49.6 & \textbf{66.1} & 59.5 & \textbf{73.5} & \textbf{84.7} & \textbf{85.7} & \textbf{58.5} & 56.1 & 61.0 & 56.7 & \textbf{64.1} & 63.6 \\
\midrule
\multicolumn{16}{c}{VLMs: 7B$-$10B} \\
\midrule
LLaVA-OV-7B & 44.0 & 52.7 & 48.4 & 57.9 & 62.8 & 66.1 & 68.4 & 74.5 & 68.4 & 61.0 & 53.7 & \textbf{70.7} & 57.8 & 60.9 & 63.4 \\
Qwen2.5-VL-7B & \textbf{60.8} & 35.2 & \textbf{60.8} & \textbf{64.2} & 67.8 & 68.3 & 70.4 & 72.4 & 78.6 & 48.8 & 56.1 & 65.9 & \textbf{61.1} & 57.9 & \textbf{68.4} \\
Qwen3-VL-8B-IT & 40.7 & 42.9 & 57.1 & 63.6 & 62.8 & \textbf{69.4} & \textbf{73.5} & 83.7 & \textbf{89.8} & \textbf{63.4} & \textbf{68.3} & 53.7 & 60.3 & 64.4 & 67.5 \\
Qwen3-VL-8B-Think & 59.3 & 54.9 & 59.3 & 57.9 & 48.8 & 65.3 & 66.3 & 74.5 & 78.6 & 61.0 & 58.5 & 65.9 & \textbf{61.1} & 59.2 & 67.3 \\
InternVL2.5-8B & 39.6 & 53.8 & 53.8 & 55.4 & 67.8 & 63.6 & 69.4 & 82.7 & 80.6 & 46.3 & 48.8 & 51.2 & 52.7 & 63.3 & 62.3 \\
InternVL3-8B & 41.8 & 47.3 & 50.5 & 57.0 & 67.8 & 68.6 & 66.3 & 81.6 & 80.6 & 43.9 & 53.7 & 53.7 & 52.3 & 62.6 & 63.4 \\
InternVL3.5-8B & 47.3 & \textbf{62.6} & 53.8 & 52.9 & 62.0 & 57.9 & 67.3 & \textbf{87.8} & 85.7 & 56.1 & 61.0 & 48.8 & 55.9 & \textbf{68.3} & 61.5 \\
InternVL3-9B & 46.2 & 56.0 & 54.9 & 61.2 & \textbf{71.9} & 65.3 & \textbf{73.5} & 82.7 & 83.7 & 48.8 & 56.1 & 51.2 & 57.4 & 66.7 & 63.8 \\
\midrule
\multicolumn{16}{c}{VLMs: $>$10B} \\
\midrule
InternVL3-14B & 52.7 & 57.1 & 63.7 & 56.2 & \textbf{77.7} & 64.5 & 80.6 & \textbf{88.8} & 90.8 & 53.7 & 48.8 & 63.4 & 60.8 & 68.1 & 70.6 \\
InternVL3.5-14B & 45.1 & 59.3 & 54.9 & 48.8 & 65.3 & 64.5 & 69.4 & 85.7 & 91.8 & 58.5 & 48.8 & 58.5 & 55.4 & 64.8 & 67.4 \\
InternVL2.5-26B & 53.8 & 62.6 & 56.0 & 60.3 & 72.7 & 71.9 & 76.5 & 84.7 & 87.8 & 56.1 & 53.7 & \textbf{68.3} & 61.7 & 68.4 & 71.0 \\
InternVL2.5-38B & 61.4 & 64.8 & 69.2 & 61.8 & 74.4 & \textbf{72.7} & \textbf{82.7} & 87.8 & \textbf{95.9} & \textbf{65.9} & \textbf{63.4} & 65.9 & \textbf{67.9} & \textbf{72.6} & \textbf{75.9} \\
InternVL3-38B & \textbf{61.5} & \textbf{65.9} & \textbf{72.5} & \textbf{66.1} & \textbf{77.7} & 66.1 & 78.6 & 85.7 & 93.9 & 63.4 & 58.5 & 58.5 & 67.4 & 72.0 & 72.8 \\
InternVL3.5-38B & 53.8 & 59.3 & 60.4 & 57.9 & 75.2 & 63.6 & 81.6 & \textbf{88.8} & 74.5 & 63.4 & \textbf{63.4} & 63.4 & 64.2 & 71.7 & 65.5 \\
\midrule
\multicolumn{16}{c}{VLMs: Proprietary} \\
\midrule
GPT-4o & 68.3 & 59.3 & 70.3 & 59.3 & 83.5 & 74.0 & 84.7 & \textbf{94.9} & \textbf{94.9} & 56.1 & 70.7 & 63.4 & 67.1 & 77.1 & 75.6 \\
GPT-5 & 67.0 & \textbf{87.9} & 68.1 & \textbf{77.7} & 84.3 & 80.2 & 85.7 & 93.9 & 91.8 & 78.0 & \textbf{78.0} & \textbf{85.4} & 77.1 & \textbf{86.0} & 81.4 \\
O1 & 62.6 & 76.9 & 68.1 & \textbf{77.7} & \textbf{89.3} & 77.7 & \textbf{90.8} & 93.9 & 93.9 & 75.6 & 68.3 & 82.9 & 76.7 & 82.1 & 80.7 \\
O3 & \textbf{76.2} & 79.1 & \textbf{74.3} & 72.4 & 86.8 & \textbf{80.5} & 86.7 & 93.9 & 93.9 & \textbf{80.5} & \textbf{78.0} & 80.5 & \textbf{79.0} & 84.5 & \textbf{82.3} \\
Gemini 2.0 Flash & 73.3 & 67.0 & 73.3 & 67.5 & 78.5 & 69.9 & 79.6 & 87.8 & 93.9 & 63.4 & 70.7 & 73.2 & 70.9 & 76.0 & 77.6 \\
Gemini 2.5 Flash & 60.4 & 68.1 & 61.5 & 72.7 & 76.0 & 71.9 & 86.7 & 91.8 & 91.8 & 70.7 & 75.6 & 78.0 & 72.7 & 77.9 & 75.8 \\
\bottomrule
\end{tabular}}
\end{table*}

\section{Blindfold Experiment on V-REX}
\label{sec:blindfold}

To investigate the extent to which V-REX relies on visual information, we conducted a ``blindfold'' experiment where models were evaluated on the benchmark without access to the input images. We evaluate the overall performance of the models on our tasks and the results are presented in Table~\ref{tab:model-category-performance}.

\begin{table*}[t]
\centering
\caption{Blindfold model overall performance on VREX tasks.}
\small
\begin{tabular}{lcccc}
\hline
Model & Guessing & Deduction & Retrieval & Navigation \\
\hline
qwen3-vl-2B-instruct & 34.71 & 21.98 & 19.51 & 55.10 \\
qwen3-vl-4B-instruct & 33.88 & 30.77 & 14.63 & 56.12 \\
qwen3-vl-8B-instruct & 24.79 & 30.77 & 7.32 & 57.14 \\
internvl3.5-1B & 37.19 & 29.67 & 34.15 & 39.80 \\
internvl3.5-2B & 19.01 & 19.78 & 21.95 & 44.90 \\
internvl3.5-4B & 29.75 & 28.57 & 9.76 & 63.27 \\
internvl3.5-8B & 12.40 & 34.07 & 14.63 & 51.02 \\
internvl3.5-14B & 30.58 & 26.37 & 24.39 & 68.37 \\
internvl3-1B & 29.75 & 30.77 & 14.63 & 43.88 \\
internvl3-2B & 25.62 & 17.58 & 17.07 & 48.98 \\
internvl3-8B & 31.40 & 30.77 & 17.07 & 62.24 \\
internvl3-14B & 25.62 & 38.46 & 31.71 & 68.37 \\
internvl3-38B & 31.40 & 35.16 & 26.83 & 71.43 \\
GPT-4o & 10.74 & 42.86 & 21.95 & 61.22 \\
GPT-5 & 36.36 & 40.66 & 29.27 & 77.55 \\
o1 & 35.54 & 35.16 & 24.39 & 74.49 \\
o3 & 33.06 & 39.56 & 31.71 & 69.39 \\
\hline
\end{tabular}
\label{tab:model-category-performance}
\end{table*}

\paragraph{Image-Critical Nature of V-REX}
The results demonstrate that for the majority of categories, specifically Guessing, Deduction, and Retrieval, model performance drops significantly in the absence of visual input. Most models achieve scores close to random guessing (e.g., around 25\% for 4-option questions or lower depending on the task format), indicating that these tasks are heavily dependent on visual understanding. The models cannot effectively solve these problems using only the textual questions and options, confirming that V-REX effectively benchmarks visual reasoning capabilities.

\paragraph{Analysis of Navigation Performance}
An notable exception is the Navigation category, where models consistently achieve higher scores even without visual context.This suggests that while visual information aids navigation, the textual components of these questions likely contain significant spatial cues or common-sense patterns that advanced language models can exploit. In this specific category, visual information may serve more as supplementary evidence rather than the sole source of critical information required for decision-making.

\section{Exploring Task Difficulty on \textit{Planning}}
\label{sec:planning_8opts}

To investigate how the size of the candidate space affects the difficulty of the \textit{Planning} task, we expand the answer space by introducing four additional distracting options for each intermediate question. This increases the number of candidate options that the model must consider during exploration process. We evaluate multiple VLMs under this setting and report the results in Table~\ref{tab:planning_8opts}.

As shown in the table, model performance consistently decreases when the number of candidate options increases. 
This trend holds across models of different sizes and architectures, indicating that larger candidate sets substantially increase the difficulty of the task. 
The performance drop suggests that models struggle more when they must select the correct reasoning step from a larger set of plausible alternatives.

Moreover, although absolute performance decreases with more options, the relative ranking of models remains largely consistent with the results obtained using fewer options. 
This indicates that the current configuration with fewer candidate options is already sufficient to approximate the comparative exploration abilities of different models, while the expanded candidate set mainly increases task difficulty without significantly altering the relative performance ordering.

\begin{table*}[t]
\centering
\caption{\textbf{Planning performance with expanded candidate space}. We increase task difficulty by expanding the candidate options for each intermediate question from $4$ to $8$. Results across Deduction, Guessing, Navigation, and Retrieval show consistent performance drops across models as the candidate space grows, indicating increased exploration difficulty. Despite this, the relative ranking of models remains largely stable.}
\label{tab:planning_8opts}
\resizebox{0.85\linewidth}{!}{\begin{tabular}{l|cc|cc|cc|cc|cc}
\toprule
& \multicolumn{2}{c|}{\textbf{Deduction}} & \multicolumn{2}{c|}{\textbf{Guessing}} & \multicolumn{2}{c|}{\textbf{Navigation}} & \multicolumn{2}{c|}{\textbf{Retrieval}} & \multicolumn{2}{c}{\textbf{Average}} \\
\cmidrule(lr){2-3}\cmidrule(lr){4-5}\cmidrule(lr){6-7}\cmidrule(lr){8-9}\cmidrule(l){10-11}
\textbf{Model} & 4 opts & 8 opts & 4 opts & 8 opts & 4 opts & 8 opts & 4 opts & 8 opts & 4 opts & 8 opts \\
\midrule
\multicolumn{11}{c}{VLMs: $<\!$7B} \\
\midrule
InternVL3-1B & 37.09 & 16.82 & 47.32 & 28.89 & 42.48 & 22.72 & 42.38 & 22.78 & 42.32 & 22.80 \\
InternVL3.5-1B & 42.36 & 20.45 & 43.48 & 19.53 & 42.25 & 20.06 & 41.11 & 22.30 & 42.30 & 20.58 \\
Qwen3-VL-2B-IT & 31.90 & 20.92 & 37.45 & 11.30 & 34.09 & 15.44 & 28.41 & 9.33 & 32.96 & 14.25 \\
InternVL3-2B & 52.73 & 32.30 & 53.84 & 29.37 & 64.09 & 30.85 & 36.87 & 21.03 & 51.88 & 28.39 \\
InternVL3.5-2B & 63.13 & 37.29 & 60.76 & 41.63 & 63.20 & 35.42 & 42.74 & 21.67 & 57.46 & 34.00 \\
Qwen3-VL-4B-IT & 62.54 & 46.03 & 65.45 & 46.32 & 70.93 & 43.02 & 36.79 & 26.59 & 58.93 & 40.49 \\
InternVL3.5-4B & 69.59 & 52.86 & 65.44 & 50.36 & 66.76 & 40.75 & 51.83 & 24.68 & 63.40 & 42.16 \\
\midrule
\multicolumn{11}{c}{VLMs: 7B--10B} \\
\midrule
Qwen3-VL-8B-IT & 74.69 & 51.19 & 84.75 & 61.41 & 79.96 & 53.68 & 46.83 & 31.31 & 71.56 & 49.40 \\
InternVL3-8B & 53.50 & 32.43 & 44.55 & 29.11 & 66.78 & 32.43 & 40.08 & 20.24 & 51.23 & 28.55 \\
InternVL3.5-8B & 66.23 & 52.91 & 69.63 & 54.26 & 73.08 & 48.50 & 47.26 & 36.87 & 64.05 & 48.13 \\
InternVL3-9B & 69.08 & 50.69 & 81.57 & 52.66 & 84.09 & 55.61 & 57.50 & 41.55 & 73.06 & 50.13 \\
\midrule
\multicolumn{11}{c}{VLMs: $>$10B} \\
\midrule
InternVL3-14B & 76.36 & 55.05 & 86.35 & 54.87 & 79.80 & 57.41 & 56.71 & 33.53 & 74.81 & 50.21 \\
InternVL3.5-14B & 78.55 & 61.44 & 73.01 & 53.84 & 77.91 & 53.20 & 53.29 & 40.36 & 70.69 & 52.21 \\
InternVL3-38B & 83.21 & 59.80 & 91.31 & 68.55 & 86.56 & 57.56 & 65.20 & 46.43 & 81.57 & 58.08 \\
InternVL3.5-38B & 81.70 & 54.52 & 85.30 & 62.01 & 84.29 & 57.30 & 57.98 & 45.67 & 77.32 & 54.88 \\
\midrule
\multicolumn{11}{c}{VLMs: Proprietary} \\
\midrule
GPT-4o & 77.92 & 47.32 & 69.61 & 36.29 & 79.65 & 46.63 & 63.69 & 34.01 & 72.72 & 41.06 \\
GPT-5 & 93.89 & 56.91 & 91.82 & 69.86 & 93.90 & 61.79 & 58.13 & 34.60 & 84.44 & 55.79 \\
O1 & 94.43 & 62.91 & 93.24 & 70.76 & 92.38 & 64.11 & 61.27 & 44.33 & 85.33 & 60.53 \\
O3 & 94.76 & 68.60 & 94.45 & 68.86 & 93.28 & 59.63 & 62.10 & 40.71 & 86.15 & 59.45 \\
\bottomrule
\end{tabular}}
\end{table*}
\section{Stepwise Recovery from Failure Analysis}
\label{appendix:recover}

\begin{figure}[h]
\centering
\includegraphics[width=0.6\linewidth]{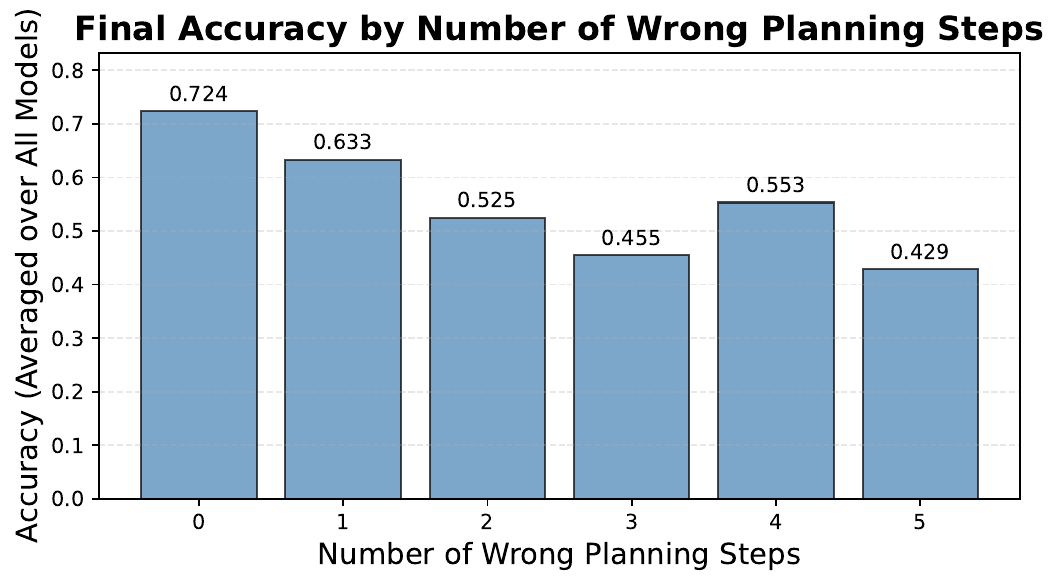}
\caption{\textbf{Stepwise result of recovery from failed planning}}
\label{fig:recover_step_planning}
\end{figure}

\begin{figure}[h]
\centering
\includegraphics[width=0.6\linewidth]{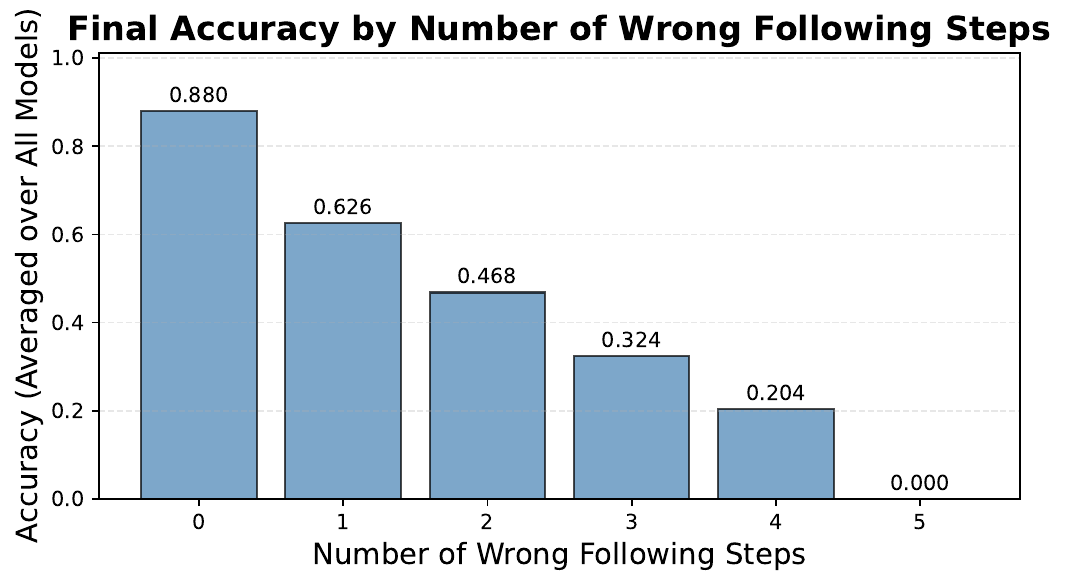}
\caption{\textbf{Stepwise result of recovery from failed following}}
\label{fig:recover_step_following}
\end{figure}

To study the model's ability to recover from failure, we investigate the relationship between the number of wrong planning or following steps in CoQ versus the final accuracy of the model, averaged over all models. As shown in Figure~\ref{fig:recover_step_planning} and Figure~\ref{fig:recover_step_following}, they display slightly different patterns. The final accuracy decreases monotonically as the number of wrong following steps increases. This reflects that more misinformation in intermediate steps would generally lead to worse results. However, the final accuracy of failed planning is not monotonic. We hypothesize that when the number of wrong planning steps is large, the model might take an entirely different approach and might still arrive at a correct conclusion. Also, the accuracy drops more sharply in Figure~\ref{fig:recover_step_following} than in Figure~\ref{fig:recover_step_planning}, which further verifies that models are generally more robust to wrong following steps than wrong planning steps.

\section{Shared Failures Analysis among Models}
\label{appendix:shared_failures}

Although several strong models (e.g. GPT-5 and O1/O3) achieve high average accuracies on Planning and Following tasks, such averages can mask \emph{process-level} failures along intermediate steps. Moreover, final-answer accuracy can be high due to recovery/shortcuts; therefore, we define sample difficulty using intermediate-step performance. To reveal whether the benchmark contains systematic difficulty (rather than isolated model-specific errors), we conduct a step-wise shared failure analysis across six model families.

We stratify samples into \textit{easy}, \textit{moderate} and \textit{hard} based on intermediate-step performance across six model families. Let a sample contain $T$ intermediate steps and let $N$ denote the number of models evaluated (namely, Qwen3-VL-8B-instruct, Gemini 2.5 Flash, GPT-5, InternVL3-9B, LLaVA-OV-7B and O1). For each step $t \in \{1,\dots,T\}$, we compute the number of models that answer step $t$ correctly. We use two sample-level statistics: (i) \textit{AvgCorrect}, mean intermediate-step accuracy across both steps and models (each model’s step outcome is binary here), and (ii) \textit{UnanimousFail}, an indicator of whether there exists at least one intermediate step that all models fail. Using these statistics, we assign difficulty as follows: \textit{Easy} if AvgCorrect $=1$; \textit{Hard} if UnanimousFail $=1$ or AvgCorrect $<0.5$; and \textit{Moderate} otherwise. Across the Following dataset (375 samples), we obtain 69 \textit{hard}, 224 \textit{moderate}, and 82 \textit{easy} samples; across the Planning dataset (351 samples), we obtain 43 \textit{hard}, 287 \textit{moderate}, and 21 \textit{easy} samples. Notably, Planning contains fewer \textit{hard} samples under our strict definition (which targets near-universal step failures), yet overall Planning accuracy can still be lower because question selection is inherently more error-prone than answering a provided step in Following.

\begin{figure}[t]
    \centering
    \includegraphics[width=\linewidth]{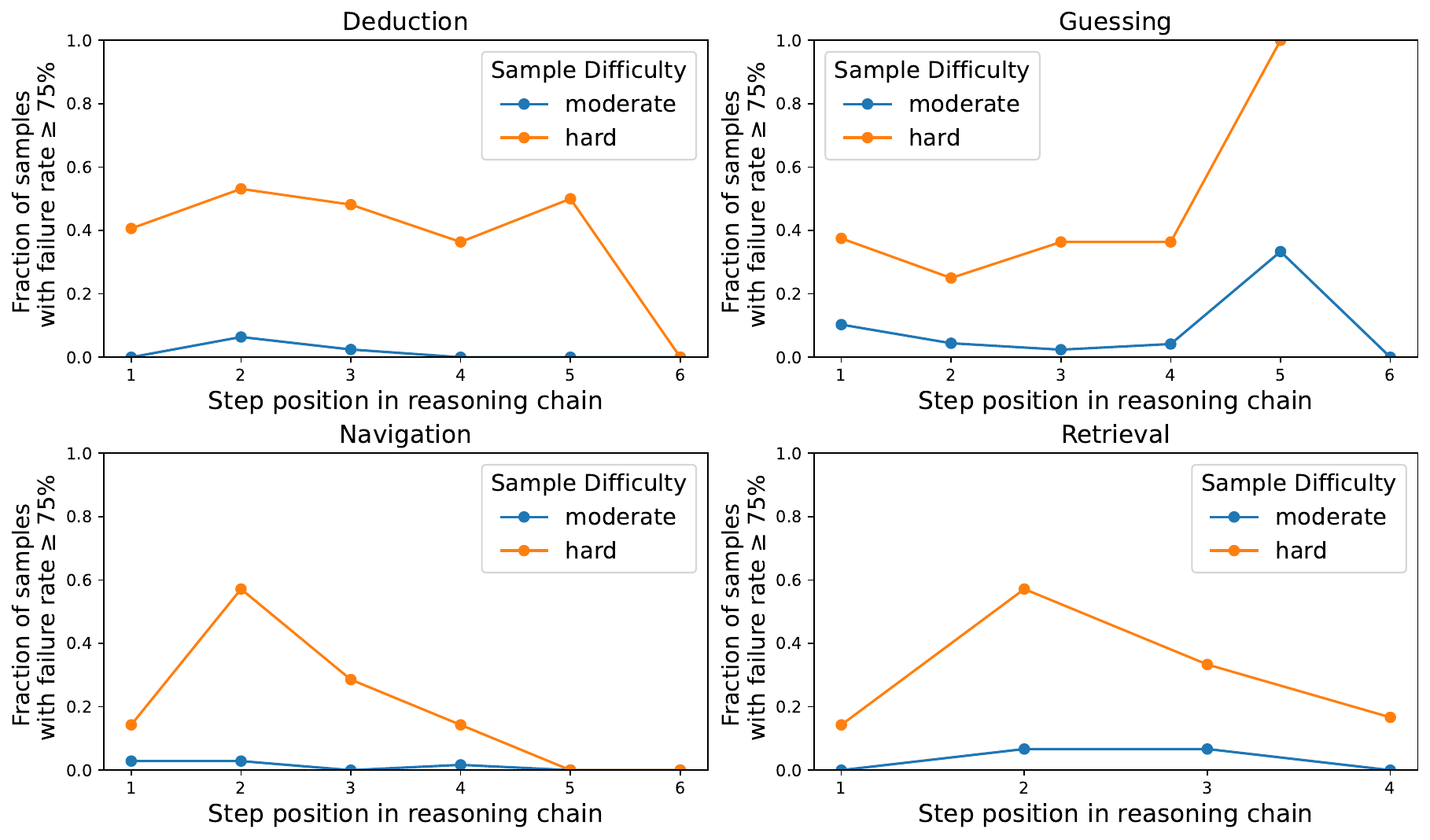}
    \caption{\textbf{Step-wise shared failure on Following (hard vs. moderate samples).} 
    For each step, we report the fraction of samples whose failure rate is at least 75\%, where the failure rate is the proportion of evaluated models that answer the step incorrectly (here $N=6$, so $\ge 75\%$ means at least $5/6$ models fail). Curves are shown separately for \textit{hard} and \textit{moderate} samples across task categories.}
    \label{fig:shared_failure_following}
\end{figure}

\begin{figure}[t]
    \centering
    \includegraphics[width=\linewidth]{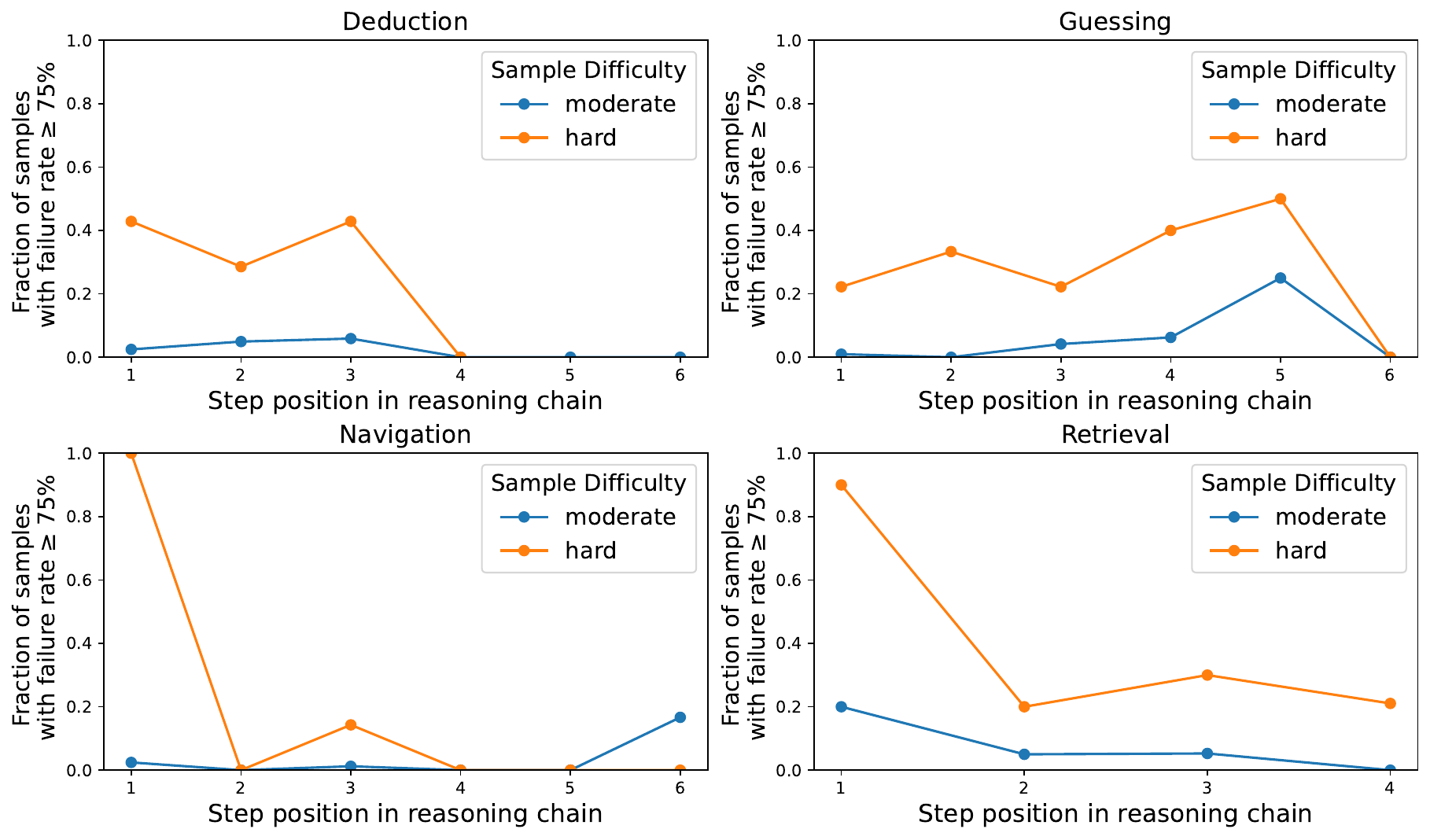}
    \caption{\textbf{Step-wise shared failure on Planning Task (hard vs. moderate samples).} 
    For each step, we report the fraction of samples whose failure rate is at least 75\%, where the failure rate is the proportion of evaluated models that answer the step incorrectly (here $N=6$, so $\ge 75\%$ means at least $5/6$ models fail). Curves are shown separately for \textit{hard} and \textit{moderate} samples across task categories.}
    \label{fig:shared_failure_planning}
\end{figure}

We then measure \emph{shared failures} step-wise by computing, at each step position, the fraction of samples for which a \emph{supermajority} of model families fail that step (fail rate $\geq 75\%$, i.e., at least 5 out of 6 models fail). Figures~\ref{fig:shared_failure_following} and \ref{fig:shared_failure_planning} compare this quantity for \textit{hard} versus \textit{moderate} samples across categories for Following and Planning tasks, respectively. While higher shared failure on \textit{hard} samples is expected given our stratification, the key observation is that failures concentrate at specific step positions: \textit{moderate} samples rarely exhibit such near-universal step failures (curves remain close to zero), whereas \textit{hard} samples show clear peaks where most model families fail simultaneously, indicating systematic bottlenecks shared across diverse model architectures. The peak locations vary by category (e.g., earlier bottlenecks in Retrieval/Navigation/Deduction and later bottlenecks in Guessing), suggesting that both early grounding and later integration/decision stages remain challenging and can propagate errors through the chain. As a result, the benchmark retains meaningful headroom even when average accuracy appears high, and it remains informative over time: progress can be measured not only by final answers but also by reductions in shared step-level failures and improved process reliability.



\section{Cases Study}
\label{appendix:cases_study}

\subsection{Success cases}
\label{appendix:success_cases}
\begin{figure*}[t]
\centering
\includegraphics[width=0.5\linewidth]{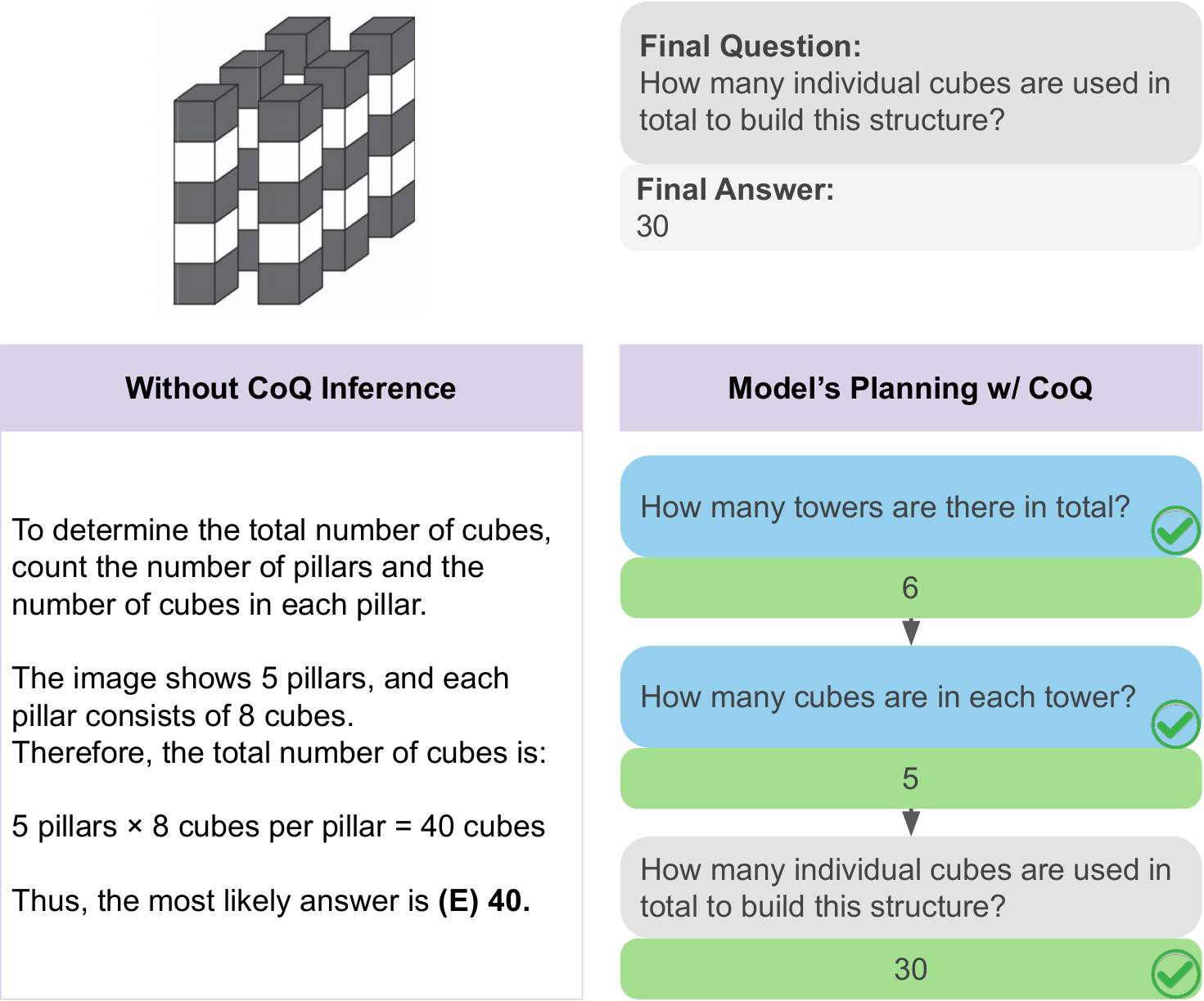}
\caption{\textbf{Success case for \textit{Planning} task.} Without CoQ, the model fails to reach the correct final answer due to perceptual and counting errors. Under the \textit{Planning} setting, however, once the model selects the correct intermediate question, it receives the corresponding answer, which compensates for its perceptual limitations and enables it to arrive at the correct final prediction.}
\label{fig:success_planning}
\vspace{-2mm}
\end{figure*}

\begin{figure*}[t]
\centering
\includegraphics[width=0.8\linewidth]{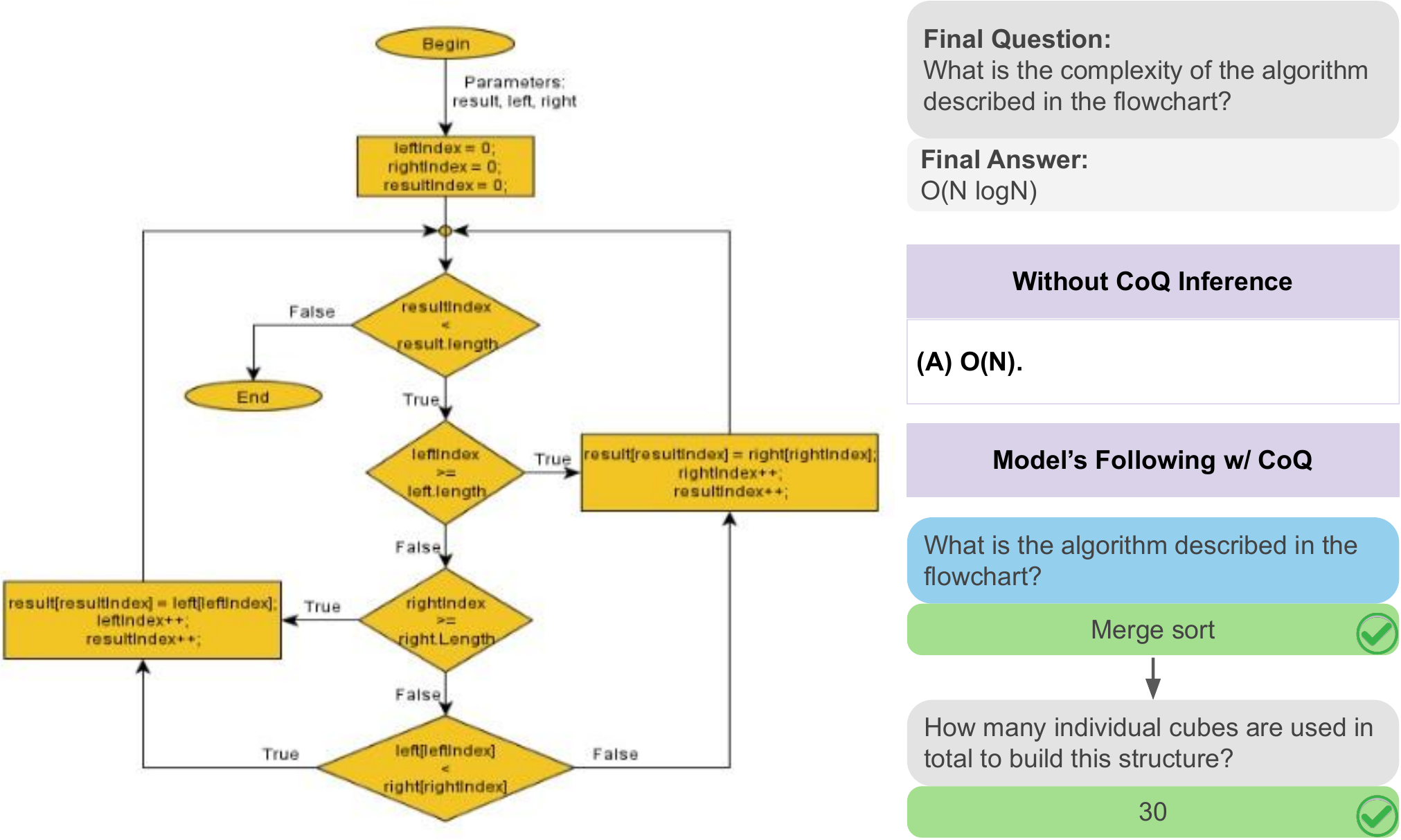}
\caption{\textbf{Success case for \textit{Following} task.} By decomposing the final question into informative sub-questions, the model correctly identifies the algorithm in the flowchart and uses this intermediate insight to arrive at the correct final answer.}
\label{fig:success_following}
\vspace{-2mm}
\end{figure*}

We showcase representative success cases enabled by CoQ across different task settings in Figure~\ref{fig:success_planning} and \ref{fig:success_following}. 
Decomposing the final question into sub-questions helps the model correct perceptual mistakes or stay aligned with the intended reasoning path, ultimately enabling it to reach the correct final answer.

\subsection{Failure cases of CoQ}
\label{appendix:failure_CoQ}
While CoQ helps organize multi-step reasoning, its advantages are limited for Retrieval category. 
These problems depend largely on straightforward factual lookup or direct visual identification, rather than layered reasoning or extended inference. 
Since Retrieval questions typically involve minimal reasoning depth, adding a CoQ chain often introduces extra steps that offer little value and may not influence the final prediction.

\begin{figure*}[t]
\centering
\includegraphics[width=0.8\linewidth]{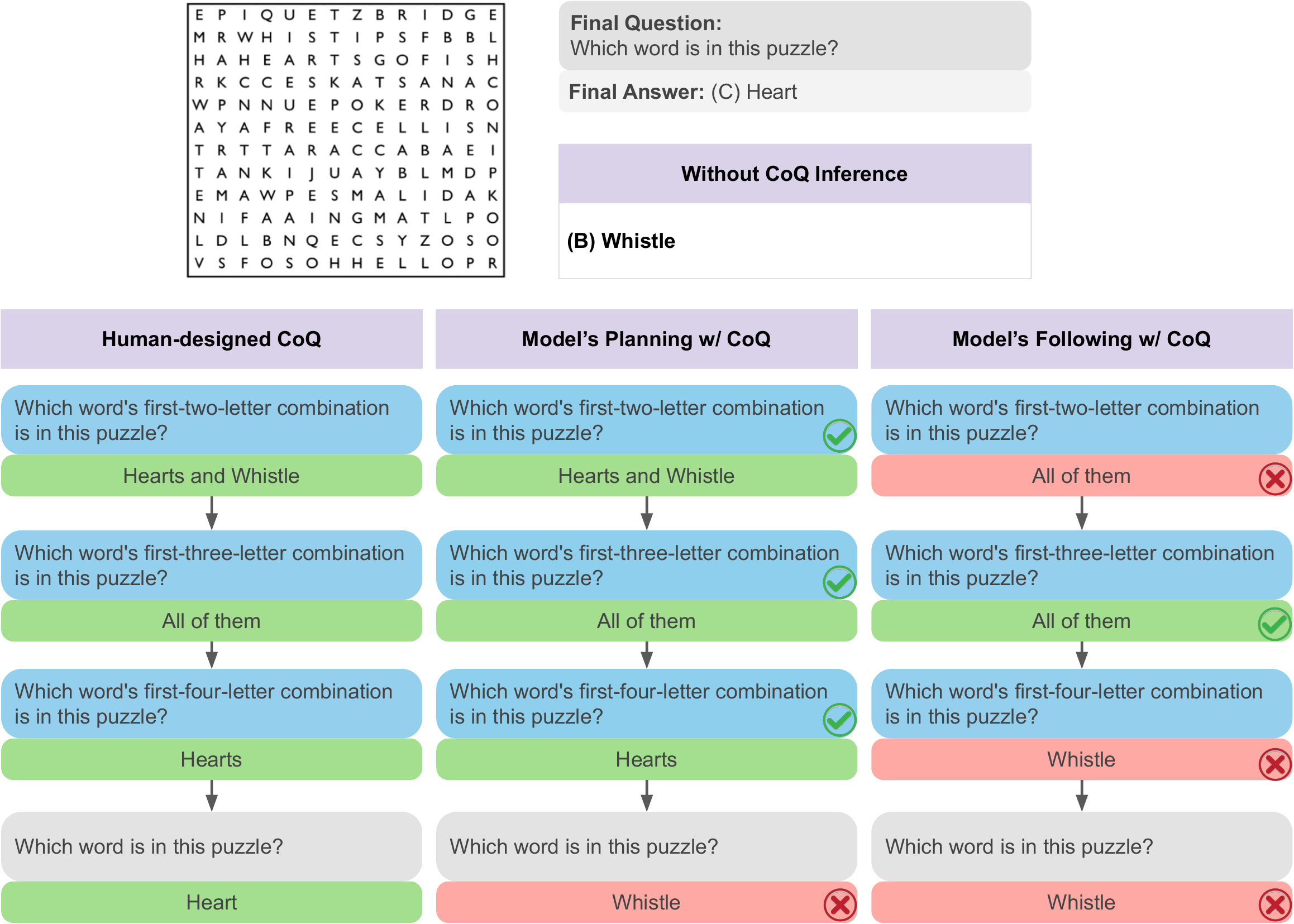}
\caption{\textbf{Failure case in the Retrieval category where CoQ provides limited benefit.}
The model can directly identify the correct word from the puzzle without CoQ. 
However, when required to follow or plan using the CoQ chain, the model becomes misled by intermediate steps and incorrectly answers “Whistle.” 
The example shows how structured CoQ exploration can interfere with tasks that primarily require precise visual matching rather than sequential reasoning.}
\label{fig:coq_failure_information}
\end{figure*}

Figure \ref{fig:coq_failure_information} illustrates a failure case where the model correctly answers the question without CoQ.
However, when the model is required to engage in step-wise exploration, it identifies the appropriate sub-questions yet still fails to reach the correct final answer.
This indicates that the bottleneck lies not in hierarchical reasoning or exploration, but in the model’s fundamental perceptual ability to extract fine-grained information from a visually dense image.

Figure \ref{fig:coq_failure_humanpath} further shows a mismatch between the human-designed reasoning path and the model’s natural reasoning trajectory. 
Under the \textit{Planning} task, the model struggles to align its question selection with the intended human chain. 
Even under the \textit{Following} task where it is required to adhere to the predefined CoQ, the model still fails to generate the correct final answer. 
This suggests that when the solution does not depend on multi-step abstraction, the CoQ structure can conflict with the model’s native retrieval process.

\begin{figure*}[t]
\centering
\includegraphics[width=0.8\linewidth]{figures/coq_failure_human_path.pdf}
\caption{\textbf{Failure case in the Retrieval category where CoQ provides limited benefit.}
Without CoQ, the model naturally checks each car one by one and correctly identifies three fully visible cars. 
Under \textit{Planning}, the imposed CoQ chain forces the model into a reasoning path that diverges from its native strategy, leading to incorrect intermediate questions and an incorrect final count. 
Even under \textit{Following}, the model cannot recover, showing that the human-designed chain is incompatible with the model’s retrieval-oriented reasoning.}
\label{fig:coq_failure_humanpath}
\end{figure*}


\subsection{Failure cases of Planning}
\label{appendix:failure_planning}

\begin{figure*}[t]
\centering
\includegraphics[width=0.7\linewidth]{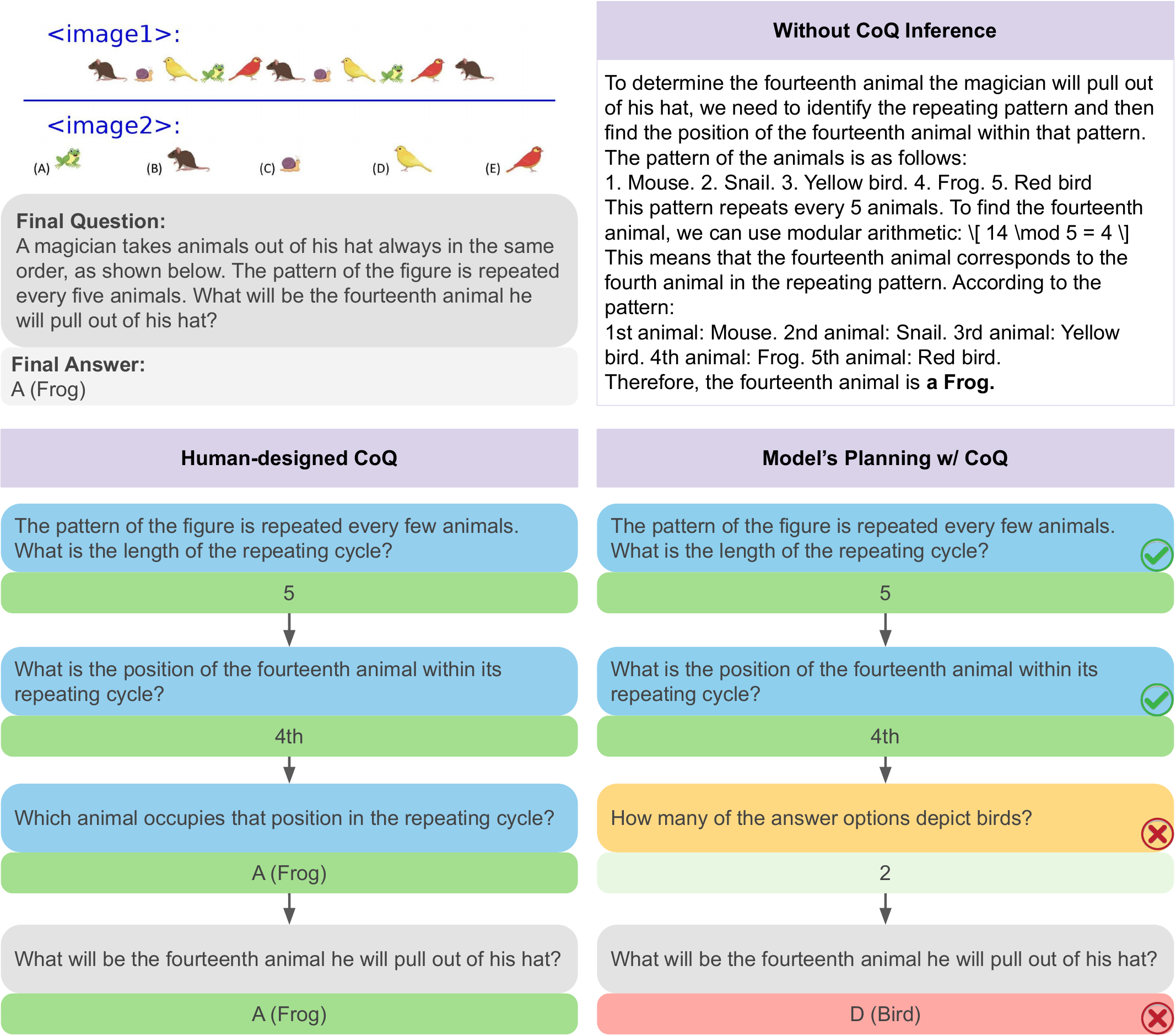}
\caption{\textbf{Failure case in the \textit{Planning} task}. 
Although the model correctly identifies the repeating cycle and the position of the fourteenth animal, its third step selects an irrelevant question about birds. 
This detour misleads the final prediction, causing the model to answer with a bird rather than the correct frog.}
\label{fig:planning_failure1}
\end{figure*}

\begin{figure*}[t]
\centering
\includegraphics[width=0.6\linewidth]{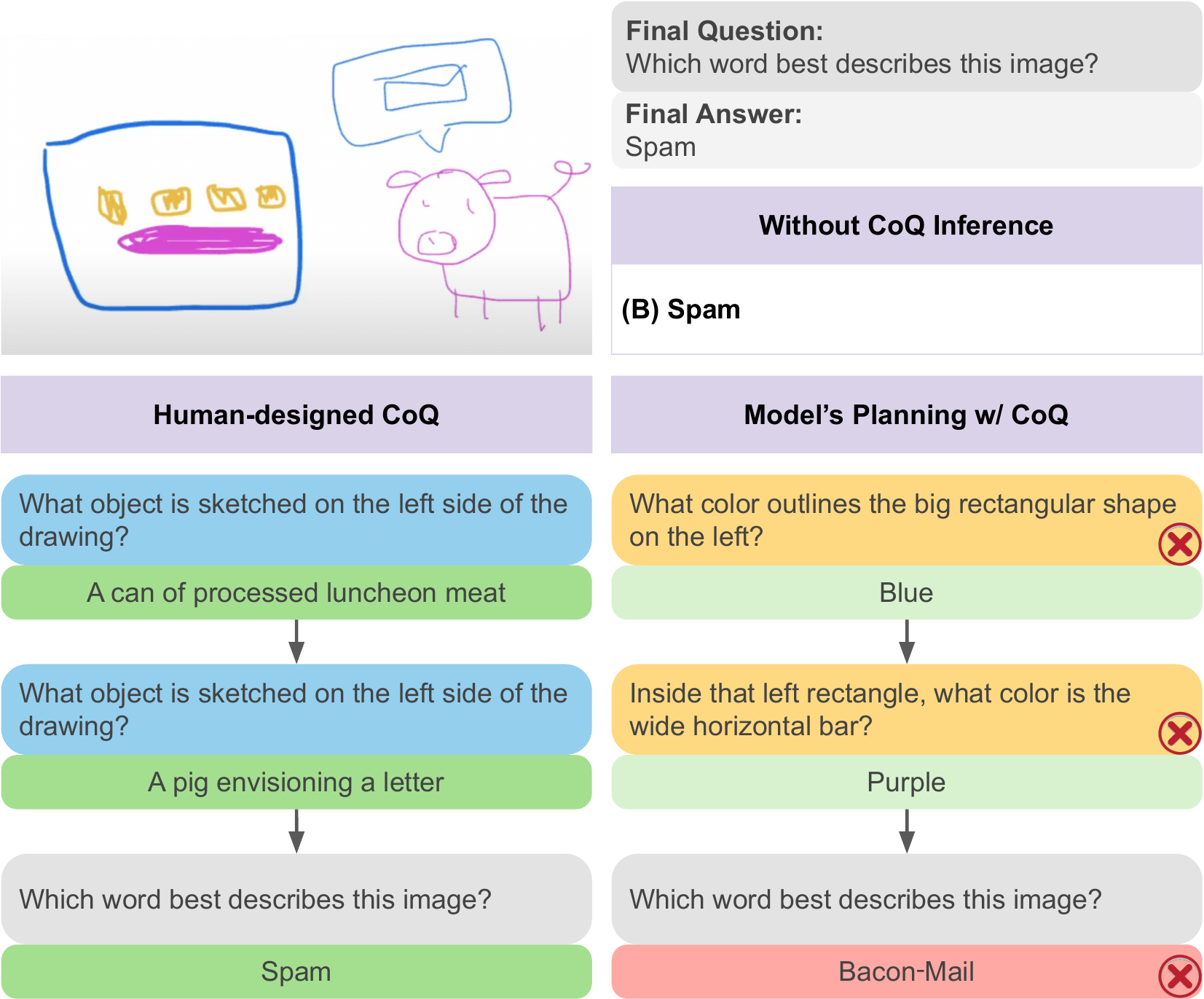}
\caption{\textbf{Failure case in the \textit{Planning} task}. 
The model selects a chain of CoQ steps centered on color descriptions, which are uninformative for determining the intended concept of the drawing. 
This distractive chain introduces additional noise and prevents the model from connecting the pig and canned meat sketches to the correct answer “Spam.”}
\label{fig:planning_failure2}
\end{figure*}

Although hierarchical question planning is intended to guide models toward more structured and interpretable reasoning, we observe that some VLMs experience performance degradation when intermediate CoQ steps are introduced. 
The expanded reasoning space can create distracting side-paths, causing models to focus on visually salient but semantically irrelevant sub-questions or to overlook key reasoning cues that are necessary for solving the task. 
In such situations, the additional steps fail to strengthen the intended chain of thought and instead introduce extra cognitive noise that disrupts both step selection and final answer prediction. 
As shown in Figure~\ref{fig:planning_failure1} and Figure~\ref{fig:planning_failure2}, models may shift their attention toward superficial objects or answer-option patterns, diverging from the human-designed chain and ultimately producing incorrect conclusions even with correct initial steps.
\looseness-1

\subsection{Failure cases of Following}
\begin{figure*}[t]
\centering
\includegraphics[width=0.6\linewidth]{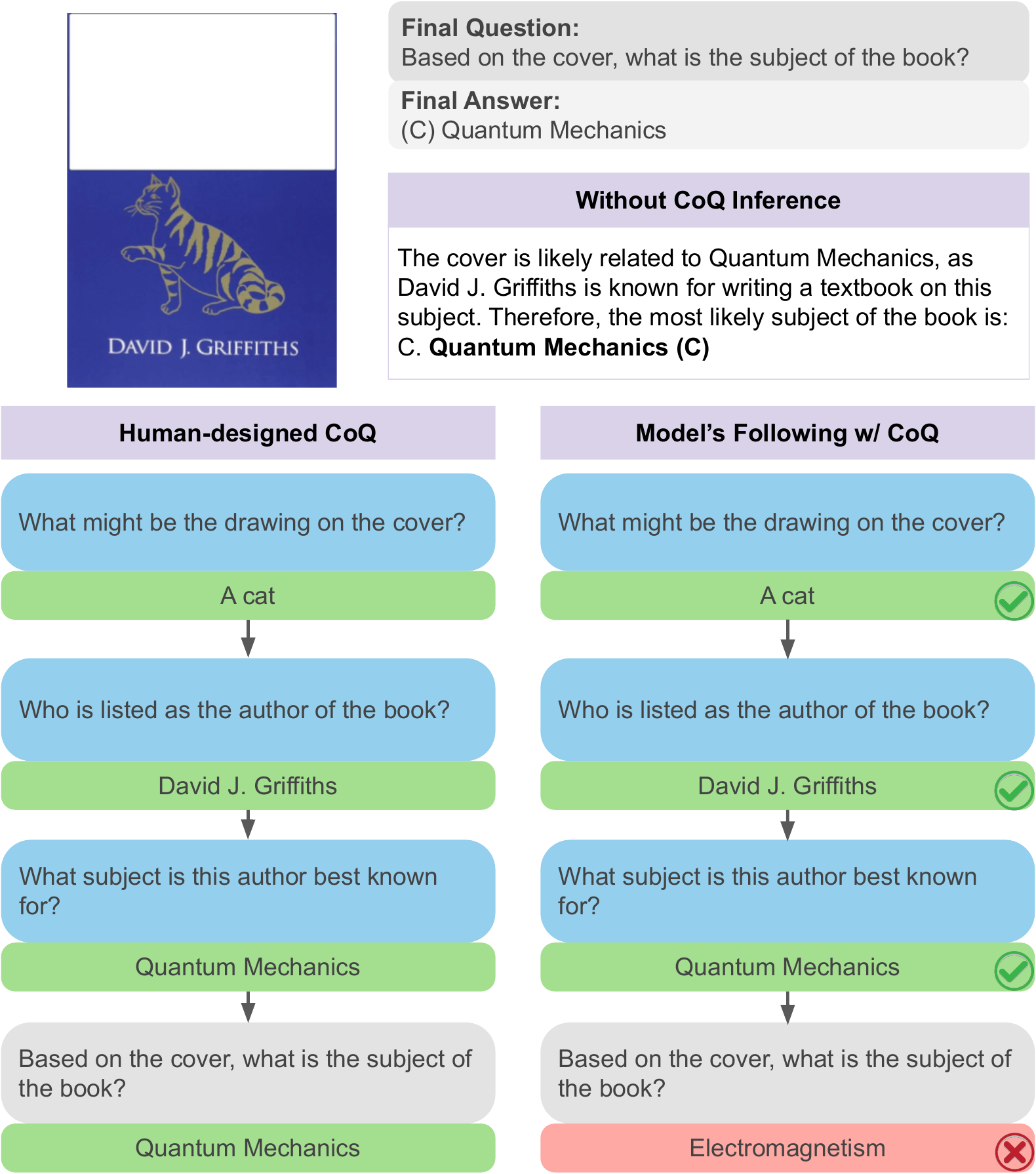}
\caption{\textbf{Failure case in the \textit{Following} task}. Although the model correctly answers all the intermediate questions about the book cover, it fails to leverage the information extracted from previous reasoning steps to answer the final question correctly.}
\label{fig:following_failure1}
\end{figure*}

\begin{figure*}[t]
\centering
\includegraphics[width=0.6\linewidth]{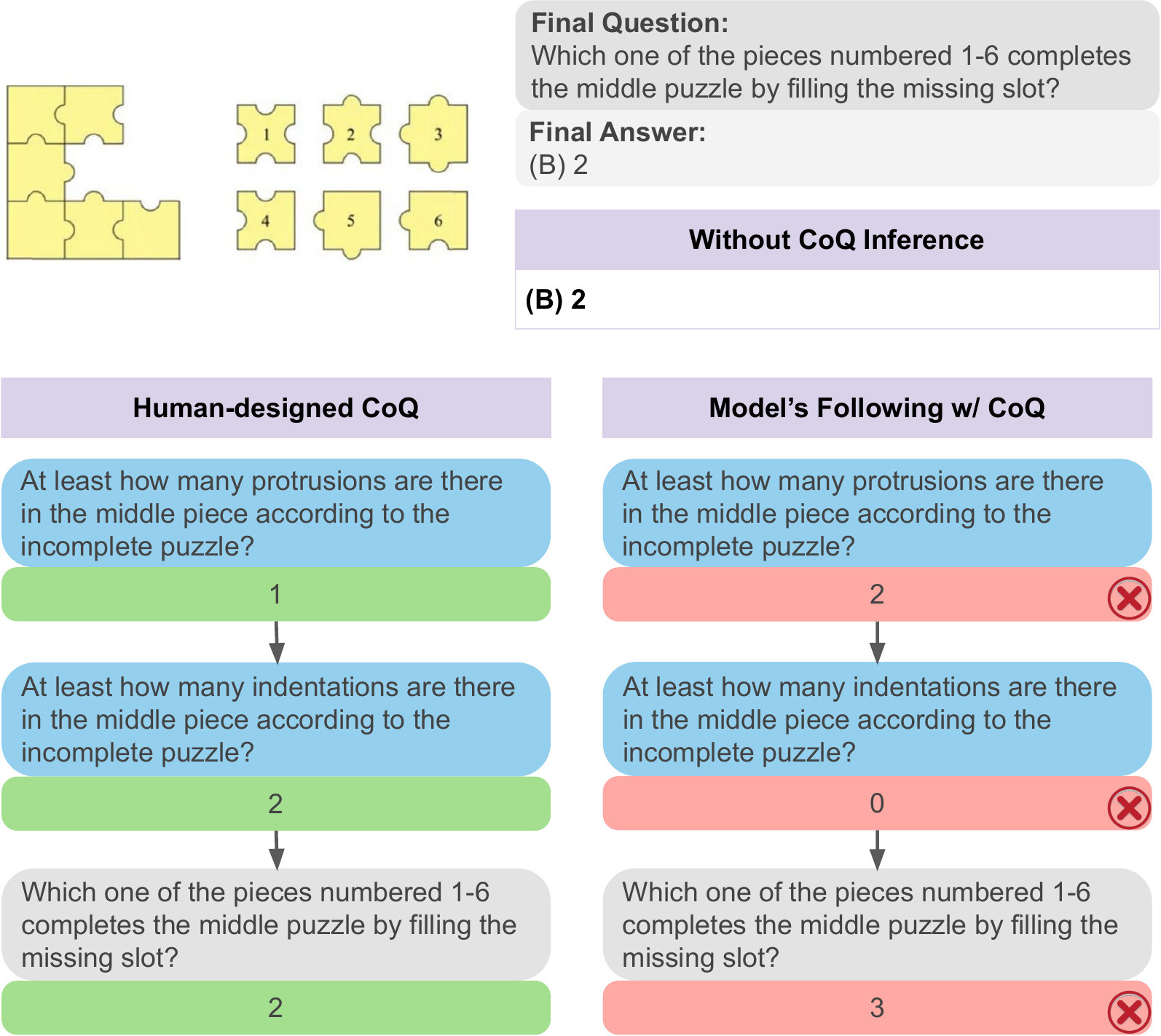}
\caption{\textbf{Failure case in the \textit{Following} task}. The model incorrectly answers the intermediate question about the least numbers of protusions and indentations of the missing puzzle piece, leading to an incorrect final answer.}
\label{fig:following_failure2}
\end{figure*}

Although the human-designed CoQ path is intended to guide the model to answer the final question with manually-crafted decomposition steps, we observe that sometimes the model would fail to answer the final question correctly after following the CoQ path. Figure~\ref{fig:following_failure1} and Figure~\ref{fig:following_failure2} show two failure cases in the \textit{Following} task that represent two typical failure modes. As shown in Figure~\ref{fig:following_failure1}, the model correctly answers all the intermediate questions about the book cover, but it fails to leverage the information extracted from previous reasoning steps to answer the final question correctly. Different from the previous case, the model in Figure~\ref{fig:following_failure2} incorrectly answers the intermediate question about the least numbers of protusions and indentations of the missing puzzle piece, leading to an incorrect final answer.

%
\end{document}